\documentclass[11pt]{amsart}
\usepackage[foot]{amsaddr}
\usepackage[
  hmarginratio={1:1},     
  vmarginratio={1:1},     
  textwidth=430pt,        
  heightrounded,          
  headsep=0.6in,          
  top=1.25in, bottom=1in, 
]{geometry}

\usepackage[utf8]{inputenc}
\usepackage[T1]{fontenc}

\usepackage{booktabs}
\usepackage{subcaption}
\usepackage[frozencache]{minted}
\usepackage{enumitem}
\setlist{noitemsep}

\usepackage{amsmath,amssymb,amsthm}
\usepackage{tikz-cd}
\tikzcdset{
  arrow style=tikz,
  diagrams={>={Straight Barb[scale=0.8]}}
}
\newcommand{\ucorner}{\rotatebox{-45}{$\ulcorner$}}

\usepackage[mathscr]{eucal}
\theoremstyle{definition}
\newtheorem*{definition}{Definition}
\newtheorem*{notation}{Notation}
\newtheorem*{remark}{Remark}

\newcommand{\cat}[1]{\mathscr{#1}}
\newcommand{\CAT}[1]{\mathbf{#1}}
\newcommand{\Cat}{\CAT{Cat}}
\newcommand{\Cart}{\CAT{Cart}}
\newcommand{\Set}{\CAT{Set}}
\newcommand{\Braid}{\sigma}
\newcommand{\Copy}{\Delta}

\newcommand{\Delete}{\lozenge}

\usepackage[backend=biber,style=authoryear,maxbibnames=5]{biblatex}
\DeclareBibliographyAlias{software}{misc}
\renewbibmacro{in:}{}
\renewcommand\cite\parencite
\renewcommand\cites\parencites
\usepackage[bookmarksnumbered,hidelinks]{hyperref}
\usepackage[capitalize,noabbrev,nosort]{cleveref}

\begin{document}

\title[Semantic enrichment of dataflow graphs]{%
  Teaching machines to understand data science code by
  semantic enrichment of dataflow graphs}

\author[E. Patterson]{Evan Patterson}
\author[I. Baldini]{Ioana Baldini}
\author[A. Mojsilovi\'c]{Aleksandra Mojsilovi\'c}
\author[K. R. Varshney]{Kush R. Varshney}

\address[Evan Patterson]{Stanford Unversity \\ Statistics Department}
\email{epatters@stanford.edu}

\address[Ioana Baldini, Aleksandra Mojsilovi\'c, Kush R. Varshney]{IBM Research AI}
\email{ioana@us.ibm.com}
\email{aleksand@us.ibm.com}
\email{krvarshn@us.ibm.com}

\begin{abstract}
  
  Your computer is continuously executing programs, but does it really
  understand them? Not in any meaningful sense. That burden falls upon human
  knowledge workers, who are increasingly asked to write and understand code.
  They deserve to have intelligent tools that reveal the connections between
  code and its subject matter. Towards this prospect, we develop an AI system
  that forms semantic representations of computer programs, using techniques
  from knowledge representation and program analysis. To create the
  representations, we introduce an algorithm for enriching dataflow graphs with
  semantic information. The semantic enrichment algorithm is undergirded by a
  new ontology language for modeling computer programs and a new ontology about
  data science, written in this language. Throughout the paper, we focus on code
  written by data scientists and we locate our work within a larger movement
  towards collaborative, open, and reproducible science.
  
\end{abstract}

\maketitle
\thispagestyle{empty} 

\section{Introduction} \label{sec:introduction}

Your computer is continuously, efficiently, and reliably executing computer
programs, but does it really understand them? Artificial intelligence
researchers have taken great strides towards teaching machines to understand
images, speech, natural text, and other media. The problem of understanding
computer code has received far less attention over the last two decades. Yet the
growth of computing's influence on society shows no signs of abating, with
knowledge workers in all domains increasingly asked to create, maintain, and
extend computer programs. For all workers, but especially those outside software
engineering roles, programming is a means to achieve practical goals, not an end
in itself. Programmers deserve intelligent tools that reveal the connections
between their code, their colleagues' code, and the subject-matter concepts to
which the code implicitly refers and to which their real enthusiasm belongs. By
teaching machines to comprehend code, we could create artificial agents that
empower human knowledge workers or perhaps even generate useful programs of
their own.

One computational domain undergoing particularly rapid growth is data science.
Besides the usual problems facing the scientist-turned-programmer, the data
scientist must contend with a proliferation of programming languages (like
Python, R, and Julia) and frameworks (too numerous to recount). Data science
therefore presents an especially compelling target for machine understanding of
computer code. An AI agent that simultaneously comprehends the generic concepts
of computing and the specialized concepts of data science could prove enormously
useful, by, for example, automatically visualizing machine learning workflows or
summarizing data analyses as natural text for human readers.

Towards this prospect, we develop an AI system that forms semantic
representations of computer programs. Our system is fully automated, inasmuch as
it expects nothing from the programmer besides the program itself and the
ability to run it. We have designed our system to handle scripts written by data
scientists, which tend to be shorter, more linear, and better defined
semantically than the large-scale codebases written by software engineers. Our
methodology is not universally applicable. Nevertheless, we think it could be
fruitfully extended to other scientific domains with a computational focus, such
as bioinformatics or computational neuroscience, by integrating it with existing
domain-specific ontologies.

We contribute several components that cohere as an AI system but also hold
independent interest. First, we define a dataflow graph representation of a
computer program, called the \emph{raw flow graph}. We extract raw flow graphs
from computer programs using static and dynamic program analysis. We define
another program representation, called the \emph{semantic flow graph}, combining
dataflow information with domain-specific information about data science. To
support the two representations, we introduce an \emph{ontology language} for
modeling computer programs, called Monocl, and an \emph{ontology} written in
this language, called the Data Science Ontology. Finally, we propose a
\emph{semantic enrichment} algorithm for transforming the raw flow graph into
the semantic flow graph. The Data Science Ontology is available
online\footnote{To browse and search the Data Science Ontology, and to see
  additional documentation, please visit
  \url{https://www.datascienceontology.org}.} and our system's source code is
available on GitHub under a permissive open source license (see
\cref{sec:conclusion}).

\subsubsection*{Organization of paper}

In the next section, we motivate our method through a pedagogical example
(\cref{sec:example}). We then explain the method itself, first informally and
with a minimum of mathematics (\cref{sec:methods}) and then again with greater
precision and rigor (\cref{sec:math}). We divide the exposition in this way
because the major ideas of the paper can be understood without the mathematical
formalism, which may be unfamiliar to some readers. We then take a step back
from technical matters to locate our work within the ongoing movement towards
collaborative, open, and reproducible data science
(\cref{sec:data-science-viewpoint}). We also demonstrate our method on a
realistic data analysis drawn from a biomedical data science challenge. In the
penultimate section, we bring out connections to existing work in artificial
intelligence, program analysis, programming language theory, and category theory
(\cref{sec:related-work}). We conclude with directions for future research and
development (\cref{sec:conclusion}). For a non-technical overview of our work,
emphasizing motivation and examples, we suggest reading
\cref{sec:introduction,sec:example,sec:data-science-viewpoint,sec:conclusion}.

\section{First examples} \label{sec:example}

We begin with a small, pedagogical example, to be revisited and elaborated
later. Three versions of a toy data analysis are shown in
\cref{lst:kmeans-scipy,lst:kmeans-sklearn,lst:kmeans-r}. The first is written in
Python using the scientific computing packages NumPy and SciPy; the second in
Python using the data science packages Pandas and Scikit-learn; and the third in
R using the R standard library. The three programs perform the same analysis:
they read the Iris dataset from a CSV file, drop the last column (labeling the
flower species), fit a $k$-means clustering model with three clusters to the
remaining columns, and return the cluster assignments and centroids.

The programs are syntactically distinct but semantically equivalent. To be more
precise, the programs are written in different programming languages---Python
and R---and the two Python programs invoke different sets of libraries.
Moreover, the programs exemplify different programming paradigms.
\cref{lst:kmeans-scipy,lst:kmeans-r} are written in functional style and
\cref{lst:kmeans-sklearn} is written in object-oriented style. Thus, at the
syntactic level, the programs appear to be very different, and conventional
program analysis tools would regard them as being very different. However, as
readers fluent in Python and R will recognize, the programs perform the same
data analysis. They are semantically equivalent, up to possibly numerical error
and minor differences in the implementation of the $k$-means clustering
algorithm. (Implementations differ mainly in how the iterative algorithm is
initialized.)

Identifying the semantic equivalence, our system furnishes the same semantic
flow graph for all three programs, shown in \cref{fig:semantic-kmeans}. The
labeled nodes and edges refer to concepts in the Data Science Ontology. The node
tagged with a question mark refers to code with unknown semantics.

\begin{figure}
  \begin{minipage}{\textwidth}
    \begin{minted}[frame=leftline,rulecolor=\color{gray!50}]{python}
import numpy as np
from scipy.cluster.vq import kmeans2

iris = np.genfromtxt('iris.csv', dtype='f8', delimiter=',', skip_header=1)
iris = np.delete(iris, 4, axis=1)

centroids, clusters = kmeans2(iris, 3)
    \end{minted}
    \captionof{listing}{$k$-means clustering in Python via NumPy and SciPy}
    \vspace{\baselineskip}
    \label{lst:kmeans-scipy}
  \end{minipage}
  \begin{minipage}[t]{0.5\textwidth}
    \begin{minted}[frame=leftline,rulecolor=\color{gray!50}]{python}
import pandas as pd
from sklearn.cluster import KMeans

iris = pd.read_csv('iris.csv')
iris = iris.drop('Species', 1)

kmeans = KMeans(n_clusters=3)
kmeans.fit(iris.values)
centroids = kmeans.cluster_centers_
clusters = kmeans.labels_
    \end{minted}
    \captionof{listing}{$k$-means clustering in Python via Pandas and Scikit-learn}
    \label{lst:kmeans-sklearn}
  \end{minipage}%
  \begin{minipage}[t]{0.5\textwidth}
    \begin{minted}[frame=leftline,rulecolor=\color{gray!50}]{r}
iris = read.csv('iris.csv',
                stringsAsFactors=FALSE)
iris = iris[, names(iris) != 'Species']

km = kmeans(iris, 3)
centroids = km$centers
clusters = km$cluster
    \end{minted}
    \captionof{listing}{$k$-means clustering in R}
    \label{lst:kmeans-r}
  \end{minipage}
\end{figure}

\begin{figure}
  \centering
  \includegraphics[width=0.6\textwidth]{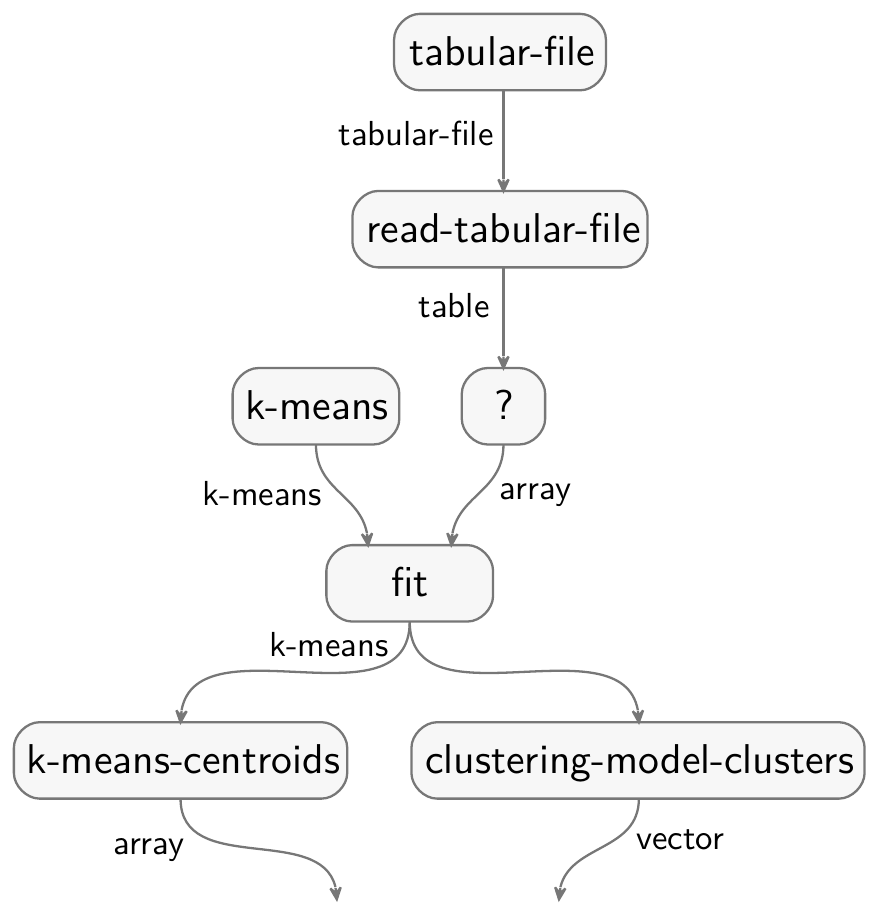}
  \caption{Semantic flow graph for three versions of $k$-means clustering
    analysis (\cref{lst:kmeans-scipy,lst:kmeans-sklearn,lst:kmeans-r})}
  \label{fig:semantic-kmeans}
\end{figure}

\section{Ideas and techniques} \label{sec:methods}

\begin{figure*}
  \centering
  \includegraphics[width=\textwidth]{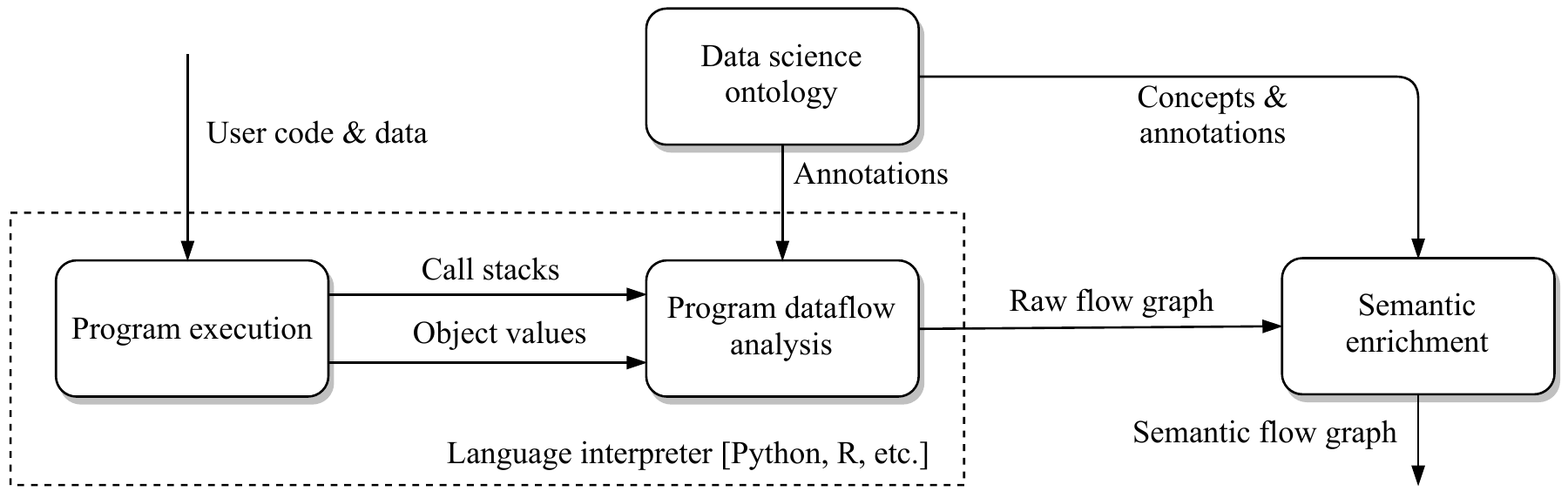}
  \caption{System architecture}
  \label{fig:architecture}
\end{figure*}

We now explain our method of constructing semantic representations of computer
programs. At the highest level, two steps connect a computer program to its
representations. First, \emph{computer program analysis} distills the raw flow
graph from the program. The \emph{raw flow graph} is a dataflow graph that
records the concrete function calls made during the execution of the program.
This graph is programming language and library dependent. In the second step, a
process of \emph{semantic enrichment} transforms the raw flow graph into the
semantic flow graph. The \emph{semantic flow graph} describes the same program
in terms of abstract concepts belonging to the Data Science Ontology. This graph
is programming language and library independent. Thus, both dataflow graphs
capture the execution of a computer program doing data analysis, but at
different levels of abstraction. The architecture diagram in
\cref{fig:architecture} summarizes our method.

Semantic enrichment requires a few supporting actors. An \emph{ontology} (or
\emph{knowledge base}), called the Data Science Ontology, underlies the semantic
content. It contains two types of knowledge: concepts and annotations.
\emph{Concepts} formalize the abstract ideas of machine learning, statistics,
and computing on data. The semantic flow graph has semantics, as its name
suggests, because its nodes and edges are linked to concepts. \emph{Annotations}
map code from data science libraries, such as Pandas and Scikit-learn, onto
concepts. During semantic enrichment, annotations determine how concrete
functions in the raw flow graph are translated into abstract functions in the
semantic flow graph.

Such, in outline, is our method. Throughout the rest of this section we develop
its elements in greater detail, beginning with the Data Science Ontology and the
ontology language in which it is expressed.

\subsection{The Data Science Ontology} \label{sec:dso}

We have begun writing an ontology---the Data Science Ontology---about
statistics, machine learning, and data processing. It aims to support automated
reasoning about data science software.

As we have said, the Data Science Ontology is comprised of concepts and
annotations. \emph{Concepts} catalog and formalize the abstract entities of data
science, such as data tables and statistical models, as well as processes that
manipulate them, like loading data from a file or fitting a model to data.
Reflecting the intuitive distinction between ``things'' and ``processes,''
concepts bifurcate into two kinds: types and functions. The terminology agrees
with that of functional programming. Thus, a \emph{type} represents a kind or
species of thing in the domain of data science. A \emph{function} is a
functional relation or mapping from an input type (the \emph{domain}) to an
output type (the \emph{codomain}). In this terminology, the concepts of a data
table and of a statistical model are types, whereas the concept of fitting a
predictive model is a function that maps an unfitted predictive model, together
with predictors and response data, to a fitted predictive model.

As a modeling assumption, we suppose that software packages for data science,
such as Pandas and Scikit-learn, concretize the concepts. \emph{Annotations} say
how this concretization occurs by mapping types and functions in software
packages onto type and function concepts in the ontology. To avoid confusion
between levels of abstraction, we call the former ``concrete'' and the latter
``abstract.'' Thus, a type annotation maps a concrete type---a primitive type or
user-defined class in a language like Python or R---onto an abstract type---a
type concept. Likewise, a function annotation maps a concrete function onto an
abstract function. We construe ``concrete function'' in the broadest possible
sense to include any programming language construct that ``does something'':
ordinary functions, methods of classes, attribute getters and setters, etc.

The division of the ontology into concepts and annotations on the one hand, and
into types and functions on the other, leads to a two-way classification.
\cref{table:ontology-classification} lists basic examples of each of the four
combinations, drawn from the Data Science Ontology.

\begin{table}
  \centering
  \caption{Example concepts and annotations from the Data Science Ontology}
  \label{table:ontology-classification}
  \begin{tabular}{lp{2in}p{2in}}
    & \textbf{Concept} & \textbf{Annotation} \\ \toprule
    \textbf{Type} &
      data table &
      pandas data frame \\ \cmidrule{2-3}
    & statistical model &
      scikit-learn estimator \\ \cmidrule{2-3}
    \textbf{Function} &
      reading a tabular data file &
      \texttt{read\_csv} function in pandas \\ \cmidrule{2-3}
    & fitting a statistical model to data &
      \texttt{fit} method of scikit-learn estimators \\ \bottomrule
  \end{tabular}
\end{table}

\begin{figure}
  \begin{subfigure}{\textwidth}
    \centering
    \includegraphics{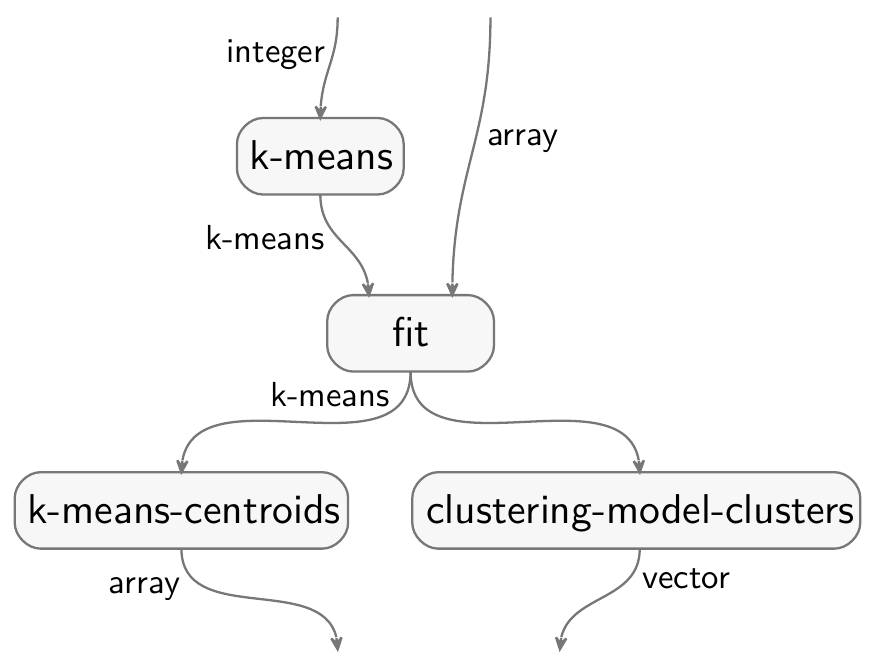}
    \caption{Annotation: \texttt{kmeans2} function in SciPy (cf.\
      \cref{lst:kmeans-scipy})}
    \label{fig:annotation-python-scipy-kmeans}
  \end{subfigure}
  \\[0.5\baselineskip]
  \begin{subfigure}{0.5\textwidth}
    \centering
    \includegraphics{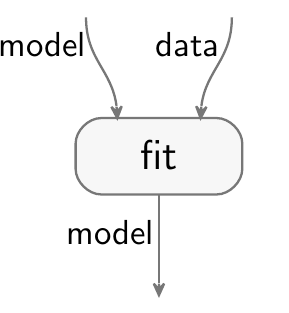}
    \caption{Annotation: \texttt{fit} method of \texttt{BaseEstimator} class in
      Scikit-learn (cf.\ \cref{lst:kmeans-sklearn})}
    \label{fig:annotation-python-sklearn-fit}
  \end{subfigure}%
  \begin{subfigure}{0.5\textwidth}
    \centering
    \includegraphics{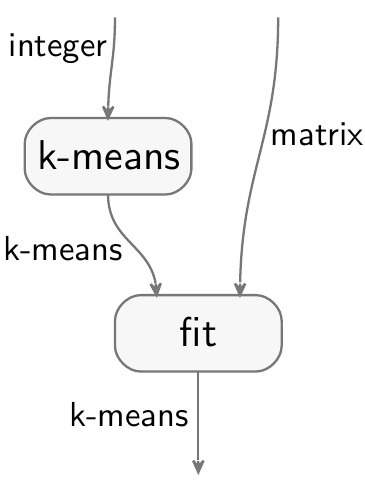}
    \caption{Annotation: \texttt{kmeans} function in R's builtin \texttt{stats}
      package (cf.\ \cref{lst:kmeans-r})}
    \label{fig:annotation-r-kmeans}
  \end{subfigure}
  \caption{Example function annotations from the Data Science Ontology}
  \label{fig:function-annotations}
\end{figure}

Significant modeling flexibility is needed to accurately translate the diverse
APIs of statistical software into a single set of universal concepts.
\cref{sec:example} shows, for example, that the concept of $k$-means clustering
can be concretized in software in many different ways. To accommodate this
diversity, we allow function annotations to map a single concrete function onto
an arbitrary abstract ``program'' comprised of function concepts. In
\cref{fig:function-annotations}, we display three function annotations relevant
to the fitting of $k$-means clustering models in
\cref{lst:kmeans-scipy,lst:kmeans-sklearn,lst:kmeans-r}. By the end of this
section, we will see how to interpret the three annotations and how the semantic
enrichment algorithm uses them to generate the semantic flow graph in
\cref{fig:semantic-kmeans}.

We have not yet said what kind of abstract ``program'' is allowed to appear in a
function annotation. Answering that question is the most important purpose of
our ontology language, to which we now turn.

\subsection{The Monocl ontology language} \label{sec:monocl}

The Data Science Ontology is expressed in an \emph{ontology language} that we
call the MONoidal Ontology and Computing Language (Monocl). We find it helpful
to think of Monocl as a minimalistic, typed, functional programming language.
The analogy usually suggests the right intuitions but is imperfect because
Monocl is simpler than any commonly used programming language, being designed
for knowledge representation rather than actual computing.

The ontology language says how to construct new types and functions from old,
for the purposes of defining concepts and annotations. Monocl is written in a
point-free textual syntax or equivalently in a graphical syntax of
interconnected boxes and wires. The two syntaxes are parallel though not quite
isomorphic. In this section, we emphasize the more intuitive graphical syntax.
We describe the constructors for types and functions and illustrate them using
the graphical syntax. A more formal development is given in
\cref{sec:concepts-as-category}.

Monocl has a minimalistic type system, supporting product and unit types as well
as a simple form of subtyping. A \emph{basic type}, sometimes called a
``primitive type,'' is a type that cannot be decomposed into simpler types.
Basic types must be explicitly defined. All other types are \emph{composite}.
For instance, the \emph{product} of two types $X$ and $Y$ is another type
$X \times Y$. It has the usual meaning: an element of type $X \times Y$ is an
element of type $X$ \emph{and} an element of type $Y$, in that order. Products
of three or more types are defined similarly. Product types are similar to
record types in conventional programming languages, such as \texttt{struct}
types in C. There is also a \emph{unit type} $1$ inhabited by a single element.
It is analogous to the \texttt{void} type in C and Java, the \texttt{NoneType}
type in Python (whose sole inhabitant is \texttt{None}), and the \texttt{NULL}
type in R.

A type can be declared a \emph{subtype} of one or more other types. To a first
approximation, subtyping establishes an ``is-a'' relationship between types. In
the Data Science Ontology, matrices are a subtype of both arrays (being arrays
of rank 2) and data tables (being tables whose columns all have the same data
type). As this example illustrates, subtyping in Monocl differs from inheritance
in a typical object-oriented programming language. Instead, subtyping should be
understood through \emph{implicit conversion}, also known as \emph{coercion}
\cite{reynolds1980,pierce1991}. The idea is that if a type $X$ is a subtype of
$X'$, then there is a canonical way to convert elements of type $X$ into
elements of type $X'$. Elaborating our example, a matrix simply \emph{is} an
array (of rank 2), hence can be trivially converted into an array. A matrix is
not strictly speaking a data table but can be converted into one (of homogeneous
data type) by assigning numerical names to the columns.

In the graphical syntax, types are represented by wires. A basic type $X$ is
drawn as a single wire labeled $X$. A product of $n$ types is a bundle of $n$
wires in parallel. The unit type is an empty bundle of wires (a blank space).
This should become clearer as we discuss wiring diagrams for functions.

A function $f$ in Monocl has an input type $X$, its \emph{domain}, and an output
type $Y$, its \emph{codomain}. We express this in the usual mathematical
notation as $f: X \to Y$. Like types, functions are either basic or composite.
Note that a basic function may have composite domain or codomain. From the
programming languages perspective, a program in the Monocl language is nothing
more than a function.

Functions are represented graphically by \emph{wiring diagrams} (also known as
\emph{string diagrams}). A basic function $f: X \to Y$ is drawn as a box labeled
$f$. The top of the box has input ports with incoming wires $X$ and the bottom
has output ports with outgoing wires $Y$. A wiring diagram defines a general
composite function by connecting boxes with wires according to certain rules.
The diagram has an outer box with input ports, defining the function's domain,
and output ports, defining the codomain.
\cref{fig:semantic-kmeans,fig:function-annotations,fig:raw-kmeans-scipy,fig:raw-kmeans-sklearn,fig:raw-kmeans-r,fig:semantic-dream-ra}
are all examples of wiring diagrams.

\newsavebox{\composebox}
\savebox{\composebox}{\includegraphics{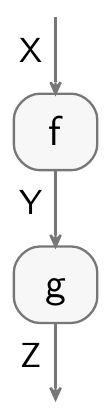}}

\begin{figure}
  \begin{subfigure}{0.25\textwidth}
    \centering
    \usebox{\composebox}
    \caption{Composition}
    \label{fig:function-composition}
  \end{subfigure}%
  \begin{subfigure}{0.25\textwidth}
    \centering
    \vbox to \ht\composebox{%
      \vfill\includegraphics{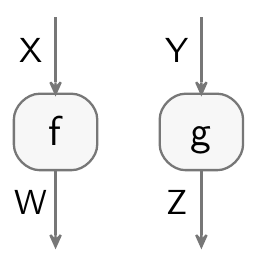}\vfill}
    \caption{Product}
    \label{fig:function-product}
  \end{subfigure}%
  \begin{subfigure}{0.25\textwidth}
    \centering
    \vbox to \ht\composebox{%
      \vfill\includegraphics{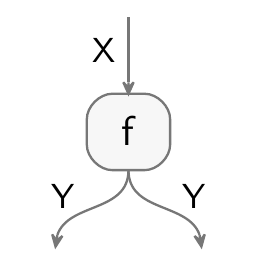}\vfill}
    \caption{Copying data}
    \label{fig:function-copy}
  \end{subfigure}%
  \begin{subfigure}{0.25\textwidth}
    \centering
    \vbox to \ht\composebox{%
      \vfill\includegraphics{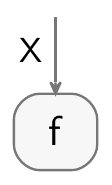}\vfill}
    \caption{Deleting data}
    \label{fig:function-delete}
  \end{subfigure}
  \caption{Graphical syntax for operations on functions}
  \label{fig:function-constructors}
\end{figure}

The rules for connecting boxes within a wiring diagram correspond to ways of
creating new functions from old. The two most fundamental ways are composing
functions and taking products of functions. The \emph{composition} of a function
$f: X \to Y$ with $g: Y \to Z$ is a new function $f \cdot g: X \to Z$, with the
usual meaning. Algorithmically speaking, $f \cdot g$ computes \emph{in
  sequence}: first $f$ and then $g$. The \emph{product} of functions $f: X \to
W$ and $g: Y \to Z$ is another function $f \times g: X \times Y \to W \times Z$.
Algorithmically, $f \times g$ computes $f$ and $g$ \emph{in parallel}, taking
the inputs, and returning the outputs, of both $f$ and $g$.
\cref{fig:function-composition,fig:function-product} show the graphical syntax
for composition and products.

The graphical syntax implicitly includes a number of special functions. For any
type $X$, the \emph{identity} function $1_X: X \to X$ maps every element of type
$X$ to itself. For each pair of types $X$ and $Y$, the \emph{braiding} or
\emph{transposition} function $\Braid_{X,Y}: X \times Y \to Y \times X$
exchanges its two inputs. Identities and braidings are drawn as straight wires
and pairs of crossed wires, respectively. Any permutation function can be
expressed by taking compositions and products of identities and braidings.
Diagrammatically, this means that a bundle of wires may be criss-crossed in
arbitrarily complex ways (provided that the wires do not bend backwards). For
each type $X$, there is also a \emph{copying} function
$\Copy_X: X \to X \times X$, which duplicates its input, and a \emph{deleting}
function $\Delete_X: X \to I$, which discards its input. In the graphical
syntax, these functions allow a single output port to have multiple or zero
outgoing wires. For instance, given a function $f: X \to Y$,
\cref{fig:function-copy,fig:function-delete} display the compositions
$f \cdot \Delta_Y: X \to Y \times Y$ and $f \cdot \Delete_Y: X \to I$. The
analogous situation is not permitted of input ports; in a well-formed wiring
diagram, every input port has exactly one incoming wire.

Besides serving as the ``is-a'' relation ubiquitous in knowledge representation
systems, the subtype relation for objects enables ad hoc polymorphism for
functions. We extend the definition of function composition to include implicit
conversion, namely, to compose a function $f: X \to Y$ with $g: Y' \to Z$, we
require not necessarily that $Y$ equals $Y'$, but only that $Y$ be a subtype of
$Y'$. Operationally, to compute $f \cdot g$, we first compute $f$, then coerce
the result from type $Y$ to $Y'$, and finally compute $g$. Diagrammatically, a
wire connecting two boxes has valid types if and only if the source port's type
is a subtype of the target port's type. Thus implicit conversions really are
implicit in the graphical syntax.

Monocl also supports ``is-a'' relations between functions, which we call
\emph{subfunctions} in analogy to subtypes. In the Data Science Ontology,
reading a table from a tabular file (call it $f$) is a subfunction of reading
data from a generic data source (call it $f'$). That sounds intuitively
plausible but what does it mean? The domain of $f$, a tabular file, is a subtype
of the domain of $f'$, a generic data source. The codomain of $f$, a table, is a
subtype of the codomain of $f'$, generic data. Now consider two possible
computational paths that take a tabular file and return generic data. We could
apply $f$, then coerce the resulting table to generic data. Alternatively, we
could coerce the tabular file to a generic data source, then apply $f'$. The
subfunction relation asserts that these two computations are equivalent. The
general definition of a subfunction is perfectly analogous.

\subsection{Raw and semantic dataflow graphs} \label{sec:graphs}

With this preparation, we can attain a more exact understanding of the raw and
semantic flow graphs. The two dataflow graphs are both wiring diagrams
representing a data analysis. However, they exist at different levels of
abstraction.

\begin{figure}
  \begin{minipage}{\textwidth}
    \centering
    \includegraphics[height=0.2\textheight]{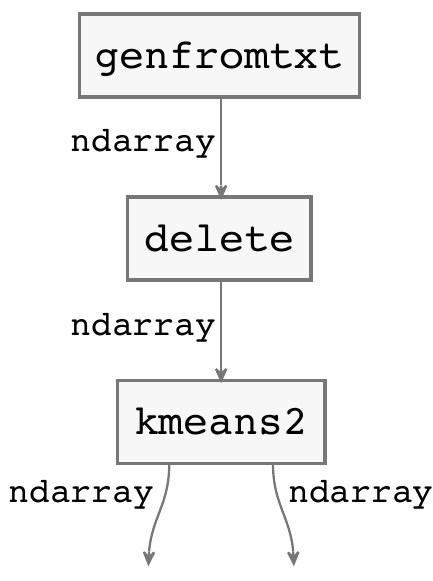}
    \caption{Raw flow graph for $k$-means clustering
      in Python via NumPy and SciPy (\cref{lst:kmeans-scipy})}
    \label{fig:raw-kmeans-scipy}
    \vspace{0.25in}
  \end{minipage}
  \begin{minipage}{0.5\textwidth}
    \centering
    \includegraphics[width=\textwidth]{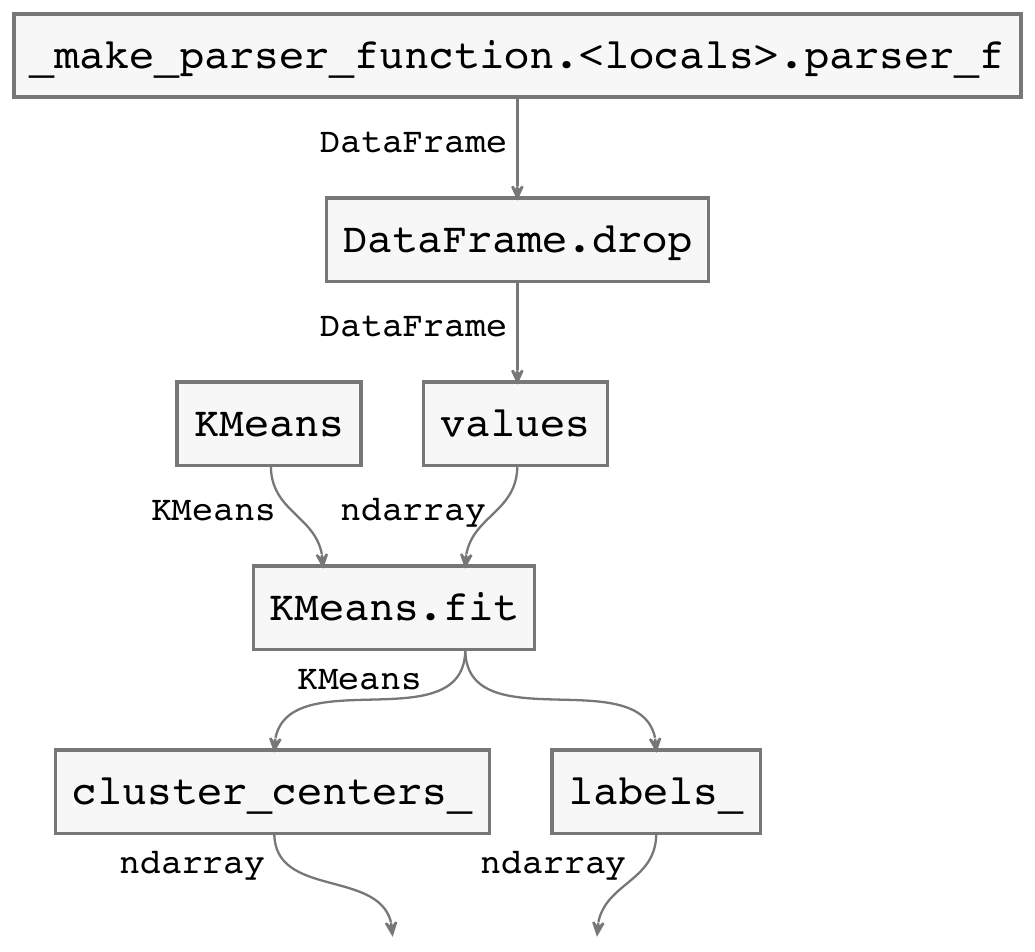}
    \caption{Raw flow graph for $k$-means clustering
      in Python via Pandas and Scikit-learn (\cref{lst:kmeans-sklearn})}
    \label{fig:raw-kmeans-sklearn}
  \end{minipage}%
  \begin{minipage}{0.5\textwidth}
    \centering
    \includegraphics[width=0.55\textwidth]{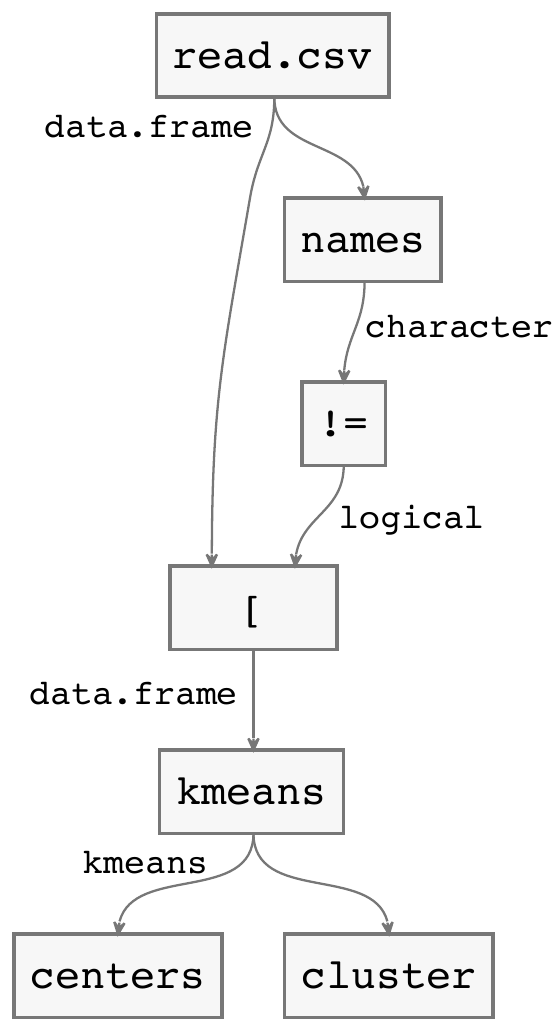}
    \caption{Raw flow graph for $k$-means clustering
      in R (\cref{lst:kmeans-r})}
    \label{fig:raw-kmeans-r}
  \end{minipage}
\end{figure}

The \emph{raw flow graph} describes the computer implementation of a data
analysis. Its boxes are concrete functions or, more precisely, the function
calls observed during the execution of the program. Its wires are concrete types
together with their observed elements. These ``elements'' are either literal
values or object references, depending on the type. To illustrate,
\cref{fig:raw-kmeans-scipy,fig:raw-kmeans-sklearn,fig:raw-kmeans-r} show the raw
flow graphs for \cref{lst:kmeans-scipy,lst:kmeans-sklearn,lst:kmeans-r},
respectively. Note that the wire elements are not shown.

The \emph{semantic flow graph} describes a data analysis in terms of universal
concepts, independent of the particular programming language and libraries used
to implement the analysis. Its boxes are function concepts. Its wires are type
concepts together with their observed elements. The semantic flow graph is thus
an abstract function, composed of the ontology's concepts and written in the
graphical syntax, but augmented with computed values. \cref{fig:semantic-kmeans}
shows the semantic flow graph for
\cref{lst:kmeans-scipy,lst:kmeans-sklearn,lst:kmeans-r}. Another semantic flow
graph is shown in \cref{fig:semantic-dream-ra} below. Again, the wire elements
are not shown.

\subsection{Program analysis} \label{sec:program-analysis}

We use computer program analysis to extract the raw flow graph from a data
analysis. Program analysis therefore plays an essential role in our AI system.
It plays an equally important role in our original publication on this topic
\cite{patterson-ibm2017}, reviewed below in \cref{sec:related-work}. In the
present work, we have extended our program analysis tools to support the R
language, but our basic methodology has changed little. We review only the major
points about our usage of computer program analysis, deferring to our original
paper for details.

Program analysis can be static or dynamic or both. \emph{Static analysis}
consumes the source code but does not execute it. Much literature on program
analysis is about static analysis because of its relevance to optimizing
compilers \cite{nielson1999,aho2006}. \emph{Dynamic analysis}, in contrast,
executes the program without necessarily inspecting the code.

Our program analysis is mainly dynamic, for a couple of reasons. Static
analysis, especially type inference, is challenging for the highly dynamic
languages popular among data scientists. Moreover, we record values computed
over the course of the program's execution, such as model parameters and
hyperparameters. For this dynamic analysis is indispensable. Of course, a
disadvantage of dynamic analysis is the necessity of running the program.
Crucially, our system needs not just the code itself, but its input data and
runtime environment. These are all requirements of scientific reproducibility
(see \cref{sec:data-science-viewpoint}), so in principle they ought to be
satisfied. In practice they are often neglected.

To build the raw flow graph, our program analysis tools record interprocedural
data flow during program execution. We begin with the empty wiring diagram and
add boxes incrementally as the program unfolds. Besides recording function calls
and their arguments and return values, the main challenge is to track the
provenance of objects as they are passed between functions. When a new box is
added to the diagram, the provenance record says how to connect the input ports
of the new box to the output ports of existing boxes.

How all this is accomplished depends on the programming language in question. In
Python, we register callbacks via \texttt{sys.settrace}, to be invoked whenever
a function is called or returns. A table of object provenance is maintained
using weak references. No modification of the abstract syntax tree (AST) is
necessary. In R, we add callbacks by rewriting the AST, to be invoked whenever a
term is (lazily) evaluated. Care must be taken to avoid breaking functions which
use nonstandard evaluation, a kind of dynamic metaprogramming unique to R
\cite{wickham2014}.

Our usage of program analysis involves conceptual as well as engineering
difficulties, because the programming model of Monocl is simpler than that of
typical programming languages. We mention just one conceptual problem to give a
sense of the issues that arise. Monocl is purely functional, whereas most
practical languages allow mutation of objects.\footnote{Mutation is more common
  in Python than in R because most R objects have copy-on-modify semantics.} Our
program analysis tools have limited capabilities for detecting mutations. When a
``function'' mutates an object, the mutated object is represented in the raw
flow graph as an extra output of the function. For instance, we interpret the
\texttt{fit} method of a Scikit-learn estimator, which modifies the model
in-place, as returning a new model (\cref{fig:annotation-python-sklearn-fit}).

\subsection{Semantic enrichment} \label{sec:semantic-enrichment}

The \emph{semantic enrichment} algorithm transforms the raw flow graph into the
semantic flow graph. It proceeds in two independent stages, one of expansion and
one of contraction. The expansion stage makes essential use of the ontology's
annotations.

\subsubsection*{Expansion}

In the \emph{expansion} stage, the annotated parts of the raw flow graph are
replaced by their abstract definitions. Each annotated box---that is, each box
referring to a concrete function annotated by the ontology---is replaced by the
corresponding abstract function. Likewise, the concrete type of each annotated
wire is replaced by the corresponding abstract type. This stage of the algorithm
is ``expansionary'' because, as we have seen, a function annotation's definition
may be an arbitrary Monocl program. In other words, a single box in the raw flow
graph may expand to an arbitrarily large subdiagram in the semantic flow graph.

The expansion procedure is \emph{functorial}, to use the jargon of category
theory. Informally, this means two things. First, notice that concrete types are
effectively annotated twice, explicitly by type annotations and implicitly by
the domain and codomain types in function annotations. Functorality requires
that these abstract types be compatible, ensuring the logical consistency of
type and function annotations. Second, expansion preserves the structure of the
ontology language, including composition and products. Put differently, the
expansion of a wiring diagram is completely determined by its action on
individual boxes (basic functions). Functorality is a modeling decision that
greatly simplifies the semantic enrichment algorithm, at the expense of imposing
restrictions on how the raw flow graph may be transformed.

\subsubsection*{Contraction}

It is practically infeasible to annotate every reusable unit of data science
source code. Most real-world data analyses use concrete types and functions that
are not annotated. This unannotated code has unknown semantics, so properly
speaking it does not belong in the semantic flow graph. On the other hand, it
usually cannot be deleted without altering the dataflow in the rest of the
diagram. Semantic enrichment must not corrupt the dataflow record.

As a compromise, in the \emph{contraction} stage, the unannotated parts of the
raw flow graph are simplified to the extent possible. All references to
unannotated types and functions are removed, leaving behind unlabeled wires and
boxes. Semantically, the unlabeled wires are interpreted as arbitrary
``unknown'' types and the unlabeled boxes as arbitrary ``unknown'' functions
(which could have known domain and codomain types). The diagram is then
simplified by \emph{encapsulating} unlabeled boxes. Specifically, every maximal
connected subdiagram of unlabeled boxes is encapsulated by a single unlabeled
box. The interpretation is that any composition of unknown functions is just
another unknown function. This stage is ``contractionary'' because it can only
decrease the number of boxes in the diagram.

\subsubsection*{Example revisited}

To reprise our original example, semantic enrichment transforms the raw flow
graphs of \cref{fig:raw-kmeans-scipy,fig:raw-kmeans-sklearn,fig:raw-kmeans-r}
into the same semantic flow graph, shown in \cref{fig:semantic-kmeans}. Let us
take a closer look at a few of the expansions and contractions involved.

Expansions related to $k$-means clustering occur in all three programs. In the
Python program based on SciPy (\cref{fig:raw-kmeans-scipy}), the
\texttt{kmeans2} function expands into a program that creates a $k$-means
clustering model, fits it to the data, and extracts its cluster assignments and
centroids, as described by the annotation in
\cref{fig:annotation-python-scipy-kmeans}. The abstract $k$-means clustering
model does \emph{not} correspond to any concrete object in the original program.
We routinely use this modeling pattern to cope with functions that are not
object-oriented with respect to models.

By contrast, the Python program based on Scikit-learn
(\cref{fig:raw-kmeans-sklearn}) is written in object-oriented style. The
\texttt{KMeans} class expands to an abstract $k$-means clustering type. The
\texttt{fit} method of the \texttt{KMeans} class is not annotated in the Data
Science Ontology. However, the \texttt{fit} method of the superclass
\texttt{BaseEstimator} \emph{is} annotated
(\cref{fig:annotation-python-sklearn-fit}), so the expansion is performed using
that annotation. As this case illustrates, subtyping and polymorphism are
indispensable when annotating object-oriented code.

The R program (\cref{fig:raw-kmeans-r}) is intermediate between these two
styles. The \texttt{kmeans} function, annotated in
\cref{fig:annotation-r-kmeans}, directly takes the data and the number of
clusters, but returns an object of class \texttt{kmeans}. The cluster
assignments and centroids are slots of this object, annotated separately. This
design pattern is typical in R, due to its informal type system.

Now consider the contractions. In the first program
(\cref{fig:raw-kmeans-scipy}), the only unannotated box is NumPy's
\texttt{delete} function. Contracting this box does not reduce the size of the
wiring diagram. A contraction involving multiple boxes occurs in the second
program (\cref{fig:raw-kmeans-sklearn}). The subdiagram consisting of the pandas
\texttt{NDFrame.drop} method composed with the \texttt{values} attribute
accessor is encapsulated into a single unlabeled box. We have left these
functions unannotated for the sake of illustration and because the section of
the Data Science Ontology dedicated to data manipulation has not yet been
developed. We expect this gap to close as the ontology grows.

\section{Mathematical foundations} \label{sec:math}

To put the foregoing ideas on a firmer footing, we formalize the ontology and
the semantic enrichment algorithm in the language of category theory. We are not
professional category theorists and we have tried to make this section
accessible to other non-category theorists. Nevertheless, readers will find it
helpful to have a working knowledge of basic category theory, as may be found in
the introductory textbooks \cite{spivak2014,awodey2010,leinster2014,riehl2016},
and of monoidal category theory, as in the survey articles
\cite{baez2010,coecke2010}. For readers without this background, or who simply
wish to understand our method informally, this section can be skipped without
loss of continuity.

Here, in outline, is our program. We take the ontology's concepts to form a
category, with type concepts corresponding to objects and function concepts
corresponding to morphisms. Defining the ontology language amounts to fixing a
categorical doctrine, which will turn out to be the doctrine of cartesian
categories with implicit conversion. Up to this point, we conform to the general
scheme of categorical knowledge representation, according to which ontologies
are simply categories in a suitable doctrine
\cite{spivak2012,patterson-arxiv2017}. Having defined the ontology's concepts as
a category $\cat C$, we then interpret the annotations as a partial functor from
a category $\cat L$ modeling a software ecosystem to the concept category $\cat
C$. Finally, we formalize the raw and semantic flow graphs as morphisms in
categories of elements over $\cat L$ and $\cat C$.

\subsection{Why category theory?} \label{sec:math-motivation}

Because category theory does not yet belong to the basic toolbox of knowledge
representation, we pause to motivate the categorical approach before launching
into the formal development. Why is category theory an appealing framework for
representing knowledge, especially about computational processes? We offer
several answers to this question.

First, there already exist whole branches of computer science, namely type
theory and programming language theory, dedicated to the mathematical modeling
of computer programs. To neglect them in knowledge representation would be
unfortunate. Category theory serves as an algebraic bridge to these fields. Due
to the close connection between category theory and type theory
\cite{crole1993,jacobs1999}---most famously, the correspondence between
cartesian closed categories and simply typed lambda theories
\cite{lambek1988}---we may dwell in the syntactically and semantically flexible
world of algebra but still draw on the highly developed theory of programming
languages. In \cref{sec:concepts-as-category}, we borrow specific notions of
subtyping and ad hoc polymorphism from programming language theory
\cite{goguen1978,reynolds1980}.

Category theory is also useful in its own right, beyond its connection to
programming language theory. The essential innovation of category theory over
the mainly syntactical theory of programming languages is that programs become
\emph{algebraic structures}, analogous to, albeit more complicated than,
classical algebraic structures like groups and monoids. Like any algebraic
structure, categories of programs are automatically endowed with an appropriate
notion of structure-preserving map between them. In this case, the
structure-preserving maps are a special kind of \emph{functor}. In
\cref{sec:annotations-as-functor}, we formulate the semantic enrichment
algorithm as a functor between categories of programs. The structuralist
philosophy underlying modern algebra is therefore central to our whole approach.

Another advantage of category theory is flexibility of syntax. Unlike the lambda
calculus and other type theories, algebraic structures like categories exist
independently of any particular system of syntax. Syntactic flexibility is
mathematically convenient but also practically important. Monoidal categories
admit a graphical syntax of \emph{wiring diagrams}, also known as \emph{string
  diagrams} \cite{baez2010,selinger2010}. We introduced the graphical syntax
informally in \cref{sec:monocl}. It offers an intuitive yet rigorous alternative
to the typed lambda calculus's textual syntax \cite{selinger2013}, which
beginners may find impenetrable. The family of graphical languages based on
string diagrams is a jewel of category theory, with applications to such diverse
fields as quantum mechanics \cite{coecke2010}, control theory \cite{baez2015a},
and natural language semantics \cite{coecke2013}.

Having arrived at the general categorical perspective, the next question to ask
is: what kind of category shall we use to model computer programs? We begin our
investigation with cartesian categories, which are perhaps the simplest possible
model of typed, functional computing. As we recall more carefully in
\cref{sec:concepts-as-category}, \emph{cartesian categories} are symmetric
monoidal categories with natural operations for copying and deleting data.
Morphisms in a cartesian category behave like mathematical functions.

As a model of computation, cartesian categories are very primitive. They do not
allow for manipulating functions as data (via lambda abstraction) or for
recursion (looping), hence they can only express terminating computations of
fixed, finite length. Extensions of this computational model abound.
\emph{Cartesian closed categories} arise as cartesian categories with a
\emph{closed} structure, whereby the whole collection of morphisms $X \to Y$ is
representable as an \emph{exponential} object $Y^X$. Closed categories have
function types, in programming jargon. According to a famous result, cartesian
closed categories are equivalent to the typed lambda calculus \cite{lambek1988}.
\emph{Traced symmetric monoidal categories} model looping and other forms of
feedback. According to another classic result, a trace on a cartesian category
is equivalent to a Conway fixed point operator \cite{hasegawa1997,hasegawa2003}.
Fixed points are used in programming language theory to define the semantics of
recursion. Combining these threads, we find in \emph{traced cartesian closed
  categories} a Turing-complete model of functional computing, amounting to the
typed lambda calculus with a fixed point operator.

Relaxing the cartesian or even the monoidal structure is another way to boost
modeling flexibility. Starting with the cartesian structure, we interpret
morphisms that are unnatural with respect to copying as performing
non-deterministic computation, such as random number generation or Monte Carlo
sampling. We interpret morphisms unnatural with respect to deleting as partial
functions, because they raise errors or are undefined on certain inputs. In a
symmetric monoidal category $\cat C$ with diagonals (not necessarily cartesian),
the morphisms that \emph{do} satisfy the naturality conditions for copying and
deleting data form a cartesian subcategory of $\cat C$, called the
\emph{cartesian center} or \emph{focus} of $\cat C$ \cite{selinger1999}. It is
also possible to relax the monoidal product itself. \emph{Symmetric premonoidal
  categories} model side effects and imperative programs, where evaluation order
matters even for parallel statements, such as variable access and assignment.
Any premonoidal category has a \emph{center} that is a monoidal category
\cite{power1997}. Thus, classical computational processes form a three-level
hierarchy: a symmetric premonoidal category has a center that is symmetric
monoidal, which in turn has a cartesian center \cite{jeffrey1997}.

This short survey hardly exhausts the categorical structures that have been used
to model computer programs. However, our purpose here is not to define the most
general model possible, but rather to adopt the \emph{simplest} model that still
captures useful information in practice. For us, that model is the cartesian
category. The structures in this categorical doctrine agree with the features
currently supported by our program analysis tools (\cref{sec:program-analysis}).
We expect that over time our software will acquire more features and achieve
better fidelity, whereupon we will adopt a more expressive doctrine. The survey
above shows that this transition can happen smoothly. In general, modularity is
a key advantage of categorical knowledge representation: category theory
provides a toolkit of mathematical structures that can be assembled in more or
less complex ways to meet different modeling needs.

\begin{notation}
  We compose our maps in diagrammatic (left-to-right) order. In particular, we
  write the composition of a morphism $f: X \to Y$ with another morphism $g: Y
  \to Z$ as $f \cdot g: X \to Z$ or simply $fg: X \to Z$. Small categories
  $\cat{C}, \cat{D}, \cat{E}, \dots$ are written in script font and large
  categories in bold font. As standard examples of the latter, we write $\Set$
  for the category of sets and functions and $\Cat$ for the category of (small)
  categories and functors. Other categories will be introduced as needed.
\end{notation}

\subsection{Concepts as category} \label{sec:concepts-as-category}

We formalize the ontology as a category. The type and function concepts in the
ontology are, respectively, the objects and morphisms that generate the
category. Abstract programs expressed in terms of concepts correspond to general
morphisms in the category, assembled from the object and morphism generators by
operations like composition and monoidal products. In this subsection, we
develop the categorical doctrine where the ontology category will reside, by
augmenting cartesian categories, motivated in \cref{sec:math-motivation}, with a
form of subtyping based on implicit conversion. Ultimately, we define a Monocl
ontology to be a finite presentation of a cartesian category with implicit
conversion.

The definition of diagonals in a monoidal category is fundamental
\cite{selinger1999}. In stating it, we take for granted the definition of a
\emph{symmetric monoidal category}; see the references at the beginning of this
section for further reading.

\begin{definition}
  A \emph{monoidal category with diagonals} is a symmetric monoidal category
  $(\cat C, \times, 1)$ together with two families of morphisms,
  \begin{equation*}
    \Copy_X: X \to X \times X \qquad\text{and}\qquad
    \Delete_X: X \to 1,
  \end{equation*}
  indexed by objects $X \in \cat C$. The morphisms $\Copy_X$ and $\Delete_X$,
  called \emph{copying} and \emph{deleting}, respectively, are required to make
  $X$ into a cocommutative comonoid (the formal dual of a commutative monoid).
  Moreover, the families must be \emph{coherent}, or \emph{uniform}, in the
  sense that $\Delete_1 = 1_1$ and for all objects $X,Y \in \cat C$, the
  diagrams commute:
  \begin{equation*}
    \begin{tikzcd}[column sep=large]
      X \times Y
        \ar{r}{\Copy_X \times \Copy_Y}
        \ar[swap]{dr}{\Copy_{X \times Y}}
      & X \times X \times Y \times Y
        \ar{d}{1_X \times \Braid_{X,Y} \times 1_Y} \\
      & X \times Y \times X \times Y
    \end{tikzcd}
    \qquad\qquad
    \begin{tikzcd}[column sep=large]
      X \times Y
        \ar{r}{\Delete_X \times \Delete_Y}
        \ar[swap]{dr}{\Delete_{X \times Y}}
      & 1 \times 1 \ar{d}{\cong} \\
      & 1
    \end{tikzcd}
  \end{equation*}
\end{definition}

As explained in graphical terms in \cref{sec:monocl}, the copying and deleting
morphisms allow data to be duplicated and discarded, a basic feature of
classical (but not quantum) computation. Uniformity is a technical condition
ensuring that copying and deleting are compatible with the symmetric monoidal
structure. So, for example, a uniform copying operation has the property that
copying data of type $X \times Y$ is equivalent to simultaneously copying data
of type $X$ and copying data of type $Y$, up to the ordering of the outputs.

A monoidal category with diagonals is a very general algebraic structure. Its
morphisms need not resemble computational processes in any conventional sense.
However, adding just one additional axiom yields the cartesian category, a
classical notion in category theory and a primitive model of functional
computing.

\begin{definition}
  A \emph{cartesian category} is a monoidal category with diagonals whose
  copying and deleting maps, $\Copy_X$ and $\Delete_X$, are \emph{natural} in
  $X$, meaning that for any morphism $f: X \to Y$, the diagrams commute:
  \begin{equation*}
    \begin{tikzcd}
      X \ar{r}{f} \ar[swap]{d}{\Copy_X} & Y \ar{d}{\Copy_Y} \\
      X \times X \ar{r}{f \times f} & Y \times Y
    \end{tikzcd}
    \qquad\qquad
    \begin{tikzcd}
      X \ar{r}{f} \ar[swap]{dr}{\Delete_X} & Y \ar{d}{\Delete_Y} \\
      & 1
    \end{tikzcd}
  \end{equation*}
  We denote by $\Cart$ the category of (small) cartesian categories and
  cartesian functors (strong monoidal functors preserving the diagonals).
\end{definition}

\begin{remark}
  Although it is not obvious, this definition of cartesian category is
  equivalent to the standard definition via the universal property of finite
  products \cite{heunen2012}. We prefer the alternative definition given here
  because it is phrased in the language of monoidal categories and string
  diagrams.
\end{remark}

In a cartesian category, the naturality conditions on copying and deleting
assert that computation is deterministic and total. In more detail, naturality
of copying says that computing a function $f$, then copying the output is the
same as copying the input, then computing $f$ on both copies. This means that
$f$ always produces the same output on a given input, i.e., $f$ is
\emph{deterministic}. Naturality of deleting says that computing the function
$f$, then deleting the output is the same as simply deleting the input. This
means that $f$ is well-defined on all its inputs, i.e., $f$ is \emph{total}.
Together, the naturality conditions establish that the category's morphisms
behave like mathematical functions.

Cartesian categories are perhaps the simplest model of typed, functional
computing, as we argued in \cref{sec:math-motivation}. We considered there
several extensions and relaxations of the cartesian structure, all centered
around morphisms. One can also entertain richer constructions on objects. In
programming jargon, this amounts to adding a more elaborate type system. A
cartesian category has a type system with product and unit types, introduced
from the programming languages perspective in \cref{sec:monocl}.

In our experience, augmenting the type system with some form of polymorphism is
a practical necessity, for the sake of code annotation and also of knowledge
representation. We will not try to summarize the large literature on
polymorphism. In keeping with the spirit of this paper, our objective is to
define the minimal practically useful system. The following definitions are
adapted, with relatively minor modifications, from Joseph Goguen and John C.\
Reynolds \cite{goguen1978,goguen1992,reynolds1980}.

\begin{definition}
  A \emph{category with implicit conversion} is a category $\cat C$ with a
  distinguished wide subcategory $\cat C_0$ containing at most one morphism
  between any two objects. If there exists a morphism $X \to X'$ in $\cat C_0$,
  we write $X \leq X'$ and say that $X$ is a \emph{subtype} of $X'$. The
  morphism $X \to X'$ itself is called an \emph{implicit conversion} or
  \emph{coercion}.
\end{definition}

\begin{remark}
  To be consistent in our usage of categorical and programming terminology, we
  ought to say that $X$ is a \emph{subobject} of $X'$. However, the term
  ``subobject'' already has an established meaning in categorical logic, which
  is related to, but different than, our usage here.
\end{remark}

We explained the informal interpretation of subtyping and implicit conversion in
\cref{sec:monocl}. One subtle point should be noted: even when types are
interpreted as sets, implicit conversions are not necessarily interpreted as set
inclusions. In the example from \cref{sec:monocl}, matrices are a subtype of
data tables, yet the set of matrices is \emph{not} a subset of the set of data
tables. (The implicit conversion function adds names to the columns of the
matrix.) Hence the slogan that ``types are not sets'' \cite{morris1973}.

Mathematically speaking, the subtype relation defines a preorder on the objects
of $\cat C$. Thus, every type $X$ is a subtype of itself. If $X$ is a subtype of
$X'$ and $X'$ a subtype of $X''$, then $X$ is a subtype of $X''$. The
corresponding implicit conversions are given by identities and by composition,
respectively. In what follows, there is no mathematical obstruction to allowing
the conversions $\cat C_0$ to form an arbitrary category, not necessarily a
preorder. That would, however, defeat the practical purpose: conversions would
need to be disambiguated by \emph{names} and hence would cease to be implicit.

When $\cat C$ is a monoidal category, we insist that implicit conversions be
compatible with the monoidal structure.

\begin{definition}
  A \emph{cartesian category with implicit conversion} is a category $\cat C$
  with implicit conversion that is also cartesian. Moreover, the implicit
  conversions $\cat C_0$ must form a \emph{monoidal} subcategory of $\cat C$.

  We denote by $\Cart_\leq$ the category whose objects are the (small) cartesian
  categories with implicit conversion and whose morphisms are the cartesian
  functors that preserve implicit conversions. For brevity, we call these
  morphisms simply ``functors.''
\end{definition}

The definition requires that subtyping be compatible with product types.
Specifically, if $X \leq X'$ and $Y \leq Y'$, then
$X \times Y \leq X' \times Y'$, with the corresponding implicit conversion given
by a product of morphisms. The subtype relation thus makes $\cat C$ into a
\emph{monoidal preorder}.

\begin{remark}
  Asking $\cat C_0$ to inherit the cartesian or even the symmetric monoidal
  structure leads to undesirable consequences, such as unwanted implicit
  conversions and even strictification of the original category $\cat C$.
  Namely, if $\cat C_0$ is a symmetric monoidal subcategory of $\cat C$, then
  the braidings $\Braid_{X,Y}: X \times Y \to Y \times X$ in $\cat C$ must
  satisfy $\Braid_{X,X} = 1_{X \times X}$, which is false under the intended
  set-theoretic interpretation.
\end{remark}

Because our notion of subtyping is operationalized by the implicit conversions,
we can extend it from objects to morphisms through naturality squares.

\begin{definition}
  Let $\cat C$ be a category with implicit conversion. A morphism $f$ in
  $\cat C$ is a \emph{submorphism} (or \emph{subfunction}) of another morphism
  $f'$, written $f \leq f'$, if in the arrow category $\cat C^\to$ there exists
  a (unique) morphism $f \to f'$ whose components are implicit conversions.

  Explicitly, if $f: X \to Y$ and $f': X' \to Y'$ are morphisms in $\cat C$,
  with $X \leq X'$ and $Y \leq Y'$, then $f \leq f'$ if and only if the diagram
  commutes:
  \begin{equation*}
    \begin{tikzcd}
      X \ar{r}{f} \ar[swap]{d}{\leq} & Y \ar{d}{\leq} \\
      X' \ar{r}{f'} & Y'
    \end{tikzcd}
  \end{equation*}
\end{definition}

\begin{remark}
  In a closed category, subtypes of basic types, $X \leq X'$ and $Y \leq Y'$,
  canonically induce subtypes of function types, $Y^{X'} \leq (Y')^X$, by
  ``restricting the domain'' and ``expanding the codomain.'' Be warned that this
  construction is \emph{not} the same as a submorphism (it is contravariant in
  $X$, while a submorphism is covariant in both $X$ and $Y$). Indeed, we do not
  treat cartesian closed categories at all in this paper.
\end{remark}

Again, see \cref{sec:monocl} for informal interpretation and examples of this
notion. Just as subtypes define a preorder on the objects of $\cat C$,
submorphisms define a preorder on the morphisms of $\cat C$. Moreover,
submorphisms respect the compositional structure of $\cat C$. They are closed
under identities, i.e., $1_X \leq 1_{X'}$ whenever $X \leq X'$, and under
composition, i.e., if $f \leq f'$ and $g \leq g'$ are composable, then
$fg \leq f'g'$. All these statements are easy to prove. To illustrate,
transitivity and closure under composition are proved by pasting commutative
squares vertically and horizontally:
\begin{equation*}
  \begin{tikzcd}
    X \ar{r}{f} \ar[swap]{d}{\leq} & Y \ar{d}{\leq} \\
    X' \ar{r}{f'} \ar[swap]{d}{\leq} & Y' \ar{d}{\leq} \\
    X'' \ar{r}{f''} & Y''
  \end{tikzcd}
  \qquad\qquad
  \begin{tikzcd}
    X \ar{r}{f} \ar[swap]{d}{\leq}
      & Y \ar{r}{g} \ar{d}{\leq}
      & Z \ar{d}{\leq} \\
    X' \ar{r}{f'} & Y' \ar{r}{g'} & Z'
  \end{tikzcd}
\end{equation*}
When $\cat C$ is a \emph{cartesian} category with implicit conversion,
submorphisms are also closed under products: if $f \leq f'$ and $g \leq g'$,
then $f \times g \leq f' \times g'$, because, by functorality, monoidal products
preserve commutative diagrams.

We now define an ontology to be nothing other than a \emph{finitely presented}
cartesian category with implicit conversion. More precisely:

\begin{definition}
  An \emph{ontology in the Monocl language} is a cartesian category with
  implicit conversion, given by a finite presentation. That is, it is the
  cartesian category with implicit conversion generated by finite sets of:
  \begin{itemize}
  \item \emph{basic types}, or \emph{object generators}, $X$
  \item \emph{basic functions}, or \emph{morphism generators}, $f: X \to Y$,
    where $X$ and $Y$ are objects
  \item \emph{basic subtypes}, or \emph{subtype generators}, $X \leq X'$, where
    $X$ and $X'$ are objects
  \item \emph{basic subfunctions}, or \emph{submorphism generators},
    $f \leq f'$, where $f: X \to Y$ and $f': X' \to Y'$ are morphisms satisfying
    $X \leq X'$ and $Y \leq Y'$
  \item \emph{function equations}, or \emph{morphism equations}, $f=g$, where
    $f,g: X \to Y$ are morphisms with equal domains and codomains.
  \end{itemize}
  If the set of morphism equations is empty, the category is called \emph{free}
  or \emph{freely generated}.
\end{definition}

Strictly speaking, a finite presentation of a category is not the same as the
category it presents. The former is a finitary object that can be represented
on, and manipulated by, a machine. The Monocl language consists of a textual and
graphical syntax for defining presentations on a computer. The latter is an
algebraic structure of infinite size, convenient for mathematical reasoning.
However, we will abuse terminology by calling both finitely presented
categories, and particular presentations thereof, ``ontologies.''

At the time of this writing, the Data Science Ontology is freely generated.
Inference in a freely generated ontology is straightforward. Deciding the
subtype or subfunction relations amounts to computing a reflexive transitive
closure. Deciding equality of objects is trivial. Deciding equality of morphisms
is the \emph{word problem} in a free cartesian category. The congruence closure
algorithm for term graphs \cite[\S 4.4]{baader1999} can be adapted to solve this
problem. In the future, the Data Science Ontology will likely include knowledge
in the form of morphism equations, creating a need for new inference procedures.
If arbitrary morphism equations are allowed, the word problem becomes
undecidable.

\subsection{Annotations as functor} \label{sec:annotations-as-functor}

If the concepts form a category, then surely the annotations ought to assemble
into a functor. Let the ontology be a cartesian category $\cat C$ with implicit
conversion. Suppose we have another such category $\cat L$, modeling a
programming language and a collection of modules written in that language. The
annotations should define a functor $F: \cat L \to \cat C$, saying how to
translate programs in $\cat L$ into programs in $\cat C$.

This tidy story does not quite survive contact with reality. We cannot expect a
finite set of formal concepts to exhaust the supply of informal concepts found
in real-world programs. Therefore any ``functor'' $F: \cat L \to \cat C$
annotating $\cat L$ must be \emph{partial}, in a sense that we will make
precise. There will be both objects and morphisms in $\cat L$ on which $F$
cannot be defined, because the category $\cat C$ is not rich enough to fully
interpret $\cat L$.

We approach partial functors indirectly, by way of partial functions. In
accordance with mathematical custom, we reduce the pre-theoretical idea of
``partial function'' to the ubiquitous notion of total function. There are two
standard ways to do this, the first based on pointed sets and the second on
spans. They are equivalent as far as sets and functions are concerned but
suggest different generalizations to categories and functors. Let us consider
them in turn.

The category of pointed sets leads to one viewpoint on partiality, popular in
programming language theory. Given a set $X$, let $X_\bot := X \sqcup \{\bot\}$
be the set $X$ with a freely adjoined base point $\bot$. A \emph{partial
  function} from $X$ to $Y$ is then a function $f: X_\bot \to Y_\bot$ preserving
the base point ($f(\bot) = \bot$). The function $f$ is regarded as ``undefined''
on the points $x \in X$ with $f(x) = \bot$. This notion of partiality can be
transported from sets to categories using enriched category theory
\cite{kelly1982,riehl2014}. Categories enriched in pointed sets, where each
hom-set has a base morphism $\bot$, have been proposed as a qualitative model of
incomplete information \cite{marsden2016}. Such categories make partiality an
all-or-nothing affair, because their composition laws satisfy $\bot \cdot f = f
\cdot \bot = \bot$ for all morphisms $f$. That is far too stringent. If we
adopted this composition law, our semantic representations would rarely be
anything but the trivial representation $\bot$.

Alternatively, a partial function can be defined as a special kind of span of
total functions. Now let us say that a \emph{partial function} from $X$ to $Y$
is a span in $\Set$
\begin{equation*}
  \begin{tikzcd}[column sep=small, row sep=small]
    & I \ar[tail,swap]{dl}{\iota} \ar{dr}{f} & \\
    X & & Y
  \end{tikzcd}
\end{equation*}
whose left leg $\iota: I \to X$ is monic (injective). The partial function's
domain of definition is $I$, which we regard as a subset of $X$. Although we
shall not need it here, we note that partial functions, and partial morphisms in
general, can be composed by taking pullbacks \cite[\S 5.5]{borceux1994c}.

We interpret the span above as \emph{partially} defining a function $f$ on $X$,
via a set of equations indexed by $I$:
\begin{equation*}
  f(x_i) := y_i, \qquad i \in I.
\end{equation*}
It is then natural to ask: what is the most general way to define a \emph{total}
function on $X$ obeying these equations? The answer is given by the pushout in
$\Set$:
\begin{equation*}
  \begin{tikzcd}[column sep=small]
    & I \ar[tail,swap]{dl}{\iota} \ar{dr}{f}
      \ar[phantom, very near end]{dd}{\ucorner} & \\
    X \ar[swap]{dr}{f_*} & & Y \ar[tail]{dl}{\iota_*} \\
    & Y_* &
  \end{tikzcd}
\end{equation*}
Because $\iota: I \to X$ is monic, so is $\iota_*: Y \to Y_*$, and we regard $Y$
as a subset of $Y_*$. The commutativity of the diagram says that $f_*$ satisfies
the set of equations indexed by $I$. The \emph{universal property} defining the
pushout says that any other function $f': X \to Y'$ satisfying the equations
factors uniquely through $f_*$, meaning that there exists a unique function $g:
Y_* \to Y'$ making the diagram commute:
\begin{equation*}
   \begin{tikzcd}
    & I \ar[swap]{dl}{\iota} \ar{dr}{f} & \\
    X \ar{r}{f_*} \ar[swap]{dr}{f'} & Y_* \ar[dashed]{d}{g} &
      Y \ar[swap]{l}{\iota_*} \ar{dl}{\iota'} \\
    & Y' &
  \end{tikzcd}
\end{equation*}
The codomain of the function $f_*: X \to Y_*$ consists of $Y$ plus a ``formal
image'' $f(x)$ for each element $x$ on which $f$ is undefined. Contrast this
with the codomain of a function $X \to Y_\bot$, which consists of $Y$ plus a
single element $\bot$ representing \emph{all} the undefined values.

This viewpoint on partiality generalizes effortlessly from $\Set$ to any
category with pushouts. We partially define the annotations as a span in
$\Cart_\leq$
\begin{equation*}
   \begin{tikzcd}[column sep=small, row sep=small]
    & \cat{I} \ar[tail,swap]{dl}{\iota} \ar{dr}{F} & \\
    \cat{L} & & \cat{C}
  \end{tikzcd}
\end{equation*}
whose left leg $\iota: \cat I \to \cat L$ is monic. We then form the pushout in
$\Cart_\leq$:
\begin{equation*}
  \begin{tikzcd}[column sep=small]
    & \cat{I} \ar[tail,swap]{dl}{\iota} \ar{dr}{F}
      \ar[phantom, very near end]{dd}{\ucorner} & \\
    \cat{L} \ar[swap]{dr}{F_*} & & \cat{C} \ar{dl}{\iota_*} \\
    & \cat{C}_* &
  \end{tikzcd}
\end{equation*}
Given a morphism $f$ in $\cat L$, which represents a concrete program, its image
$F_*(f)$ in $\cat{C}_*$ is a partial translation of the program into the
language defined by the ontology's concepts.

The universal property of the pushout in $\Cart_\leq$, stated above in the case
of $\Set$, gives an appealing intuitive interpretation to program translation.
The category $\cat C$ is not rich enough to fully translate $\cat L$ via a
functor $\cat L \to \cat C$. As a modeling assumption, we suppose that $\cat C$
has some ``completion'' $\overline{\cat C}$ for which a full translation
$\overline F: \cat L \to \overline{\cat C}$ \emph{is} possible. We do not know
$\overline{\cat C}$, or at the very least we cannot feasibly write it down.
However, if we take the pushout functor $F_*: \cat{L} \to \cat{C}_*$, we can at
least guarantee that, no matter what the complete translation $\overline{F}$ is,
it will factor through $F_*$. Thus $F_*$ defines the most general possible
translation, given the available information.

The properties of partial functions largely carry over to partial functors, with
one important exception: the ``inclusion'' functor $\iota_*: \cat{C} \to
\cat{C}_*$ need not be monic, even though $\iota: \cat{I} \to \cat{L}$ is.
Closely related is the fact that $\Cart_\leq$ (like its cousins $\Cat$ and
$\Cart$, but unlike $\Set$) does not satisfy the \emph{amalgamation property}
\cite{macdonald2009}. To see how $\iota_*$ can fail to be monic, suppose that
the equation $f_1 \cdot f_2 = f_3$ holds in $\cat{L}$ and that the defining
equations include $F(f_i) := g_i$ for $i=1,2,3$. Then, by the functorality of
$F_*$, we must have $g_1 \cdot g_2 = g_3$ in $\cat{C}_*$, even if $g_1 \cdot g_2
\neq g_3$ in $\cat{C}$. Thus the existence of $F_*$ can force equations between
morphisms in $\cat{C}_*$ that do not hold in $\cat{C}$.

When the categories in question are finitely presented, the pushout functor also
admits a finitary, equational presentation, suitable for computer algebra. Just
as we define an ontology to be a finitely presented category, we define an
ontology with annotations to be a finitely presented functor.

\begin{definition}
  An \emph{ontology with annotations in the Monocl language} is a functor
  between cartesian categories with implicit conversion, defined by a finite
  presentation. Explicitly, it is generated by:
  \begin{itemize}
  \item a finite presentation of a category $\cat C$ in $\Cart_\leq$, the
    ontology category;
  \item a finite presentation of a category $\cat L$ in $\Cart_\leq$, the
    programming language category; and
  \item a finite set of equations partially defining a functor $F$ from $\cat L$
    to $\cat C$.
  \end{itemize}
\end{definition}

The equations partially defining the functor $F$ may be indexed by a category
$\cat I$, in which case they take the form
\begin{equation*}
  F(X_i) := Y_i, \qquad
    X_i \in \cat L, \quad
    Y_i \in \cat C,
\end{equation*}
for each $i \in \cat I$, and
\begin{equation*}
  F(f_k) := g_k, \qquad
    f_k \in \cat L(X_i,X_j), \quad
    g_k \in \cat C(Y_i,Y_j),
\end{equation*}
for each $i,j \in \cat I$ and $k \in \cat I(i,j)$. The equations present a span
$\cat{L} \overset{\iota}\leftarrowtail \cat{I} \overset{F}\rightarrow \cat{C}$
whose left leg is monic and the functor generated by the equations is the
pushout functor $F_*: \cat{L} \to \cat{C}_*$ described above.

\begin{remark}
  Our two definitions involving finite presentations are not completely
  rigorous, but can be made so using generalized algebraic theories
  \cite{cartmell1978,cartmell1986}. There is a generalized algebraic theory of
  cartesian categories with implicit conversion, whose category of models is
  $\Cart_\leq$, and a theory of functors between them, whose category of models
  is the arrow category $\Cart_\leq^\to$. Cartmell gives as simpler examples the
  theory of categories, with models $\Cat$, and the theory of functors, with
  models $\Cat^\to$ \cite{cartmell1986}. Any category of models of a generalized
  algebraic theory is cocomplete and admits free models defined by finite
  presentations.
\end{remark}

Before closing this subsection, we should acknowledge what we have left
unformalized. In construing the annotations as a functor, we model programming
languages like Python and R as cartesian categories with implicit conversion. We
do not attempt to do so rigorously. The formal semantics of Python and R are
quite intricate and exist only in fragments \cite{guth2013,morandat2012}. Our
program analysis involves numerous simplifications, infidelities, and
heuristics, as sketched in \cref{sec:program-analysis}. Even if we could
complete it, a formalization would probably be too complicated to illuminate
anything about our method. We thus rest content with an informal understanding
of the relationship between Monocl and full-fledged programming languages like
Python and R.

\subsection{Flow graphs and categories of elements}

To a first approximation, the raw and semantic flow graphs are morphisms in the
categories $\cat L$ and $\cat C_*$, respectively. The expansion stage of the
semantic enrichment algorithm simply applies the annotation functor
$F_*: \cat L \to \cat C_*$ to a morphism in $\cat L$. The contraction stage, a
purely syntactic operation, groups together morphisms in $\cat C_*$ that are not
images of $\cat C$ under the inclusion functor $\iota_*: \cat C \to \cat C_*$.

To complete the formalization of semantic enrichment, we must account for the
observed elements in the raw and semantic flow graphs. As noted in
\cref{sec:methods}, flow graphs capture not only the types and functions
comprising a program, but also the values computed by the program. In category
theory, values can be bundled together with objects and morphisms using a device
known as the \emph{category of elements}. We formalize the raw and semantic flow
graphs as morphisms in suitable categories of elements.

The objects and morphisms in the ontology category $\cat C$ can be, in
principle, interpreted as sets and functions. A set-theoretic
\emph{interpretation} of $\cat C$ is a cartesian functor
$I_{\cat C}: \cat C \to \Set$. In programming language terms, $I_{\cat C}$ is a
\emph{denotational semantics} for $\cat C$. Suppose the concrete language
$\cat L$ also has an interpretation $I_{\cat L}: \cat L \to \Set$. Assuming the
equations partially defining the annotation functor are true under the
set-theoretic interpretations, the diagram below commutes:
\begin{equation*}
  \begin{tikzcd}[column sep=small]
    & \cat{I} \ar[tail,swap]{dl}{\iota} \ar{dr}{F} & \\
    \cat{L} \ar[swap]{dr}{I_{\cat L}} & & \cat{C} \ar{dl}{I_{\cat C}} \\
    & \Set &
  \end{tikzcd}
\end{equation*}
By the universal property of the annotation functor $F_*$, there exists a unique
interpretation $I_{\cat C_*}: \cat C_* \to \Set$ making the diagram commute:
\begin{equation*}
  \begin{tikzcd}
    \cat{L} \ar{r}{F_*} \ar[swap]{dr}{I_{\cat L}}
      & \cat{C}_* \ar[dashed,pos=0.33]{d}{I_{\cat{C}_*}}
      & \cat{C} \ar[swap]{l}{\iota_*} \ar{dl}{I_{\cat C}} \\
    & \Set &
  \end{tikzcd}
\end{equation*}

Each of these three interpretations yields a category of elements, also known as
a ``Grothendieck construction'' \cites[\S 12.2]{barr1990}[\S 2.4]{riehl2016}.
\begin{definition}
  The \emph{category of elements} of a cartesian functor $I: \cat C \to \Set$
  has as objects, the pairs $(X,x)$, where $X \in \cat C$ and $x \in I(X)$, and
  as morphisms $(X,x) \to (Y,y)$, the morphisms $f: X \to Y$ in $\cat C$
  satisfying $I(f)(x) = y$.
\end{definition}

The category of elements of a cartesian functor $I: \cat C \to \Set$ is itself a
cartesian category. Composition and identities are inherited from $\cat C$.
Products are defined on objects by
\begin{equation*}
  (X,x) \times (Y,y) := (X \times Y, (x,y))
\end{equation*}
and on morphisms exactly as in $\cat C$, and the unit object is $(1,*)$, where
$*$ is an arbitrary fixed element. The diagonals are also inherited from
$\cat C$, taking the form
\begin{equation*}
  \Copy_{(X,x)}: (X,x) \to (X \times X, (x,x)), \qquad
  \Delete_{(X,x)}: (X,x) \to (1,*).
\end{equation*}

We may at last define a \emph{raw flow graph} to be a morphism in the category
of elements of $I_{\cat L}$. Likewise, a \emph{semantic flow graph} is a
morphism in the category of elements of $I_{\cat C_*}$. Note that the
interpretations of $\cat L$, $\cat C$, and $\cat C_*$ are conceptual devices; we
do not actually construct a denotational semantics for the language $\cat L$ or
the ontology $\cat C$. Instead, the program analysis tools observe a
\emph{single} computation and produce a \emph{single} morphism $f$ in the
category of elements of $I_{\cat L}$. By the definition of the interpretation
$I_{\cat C_*}$, applying the annotation functor $F_*: \cat L \to \cat C_*$ to
this morphism $f$ yields a morphism $F_*(f)$ belonging to the category of
elements of $I_{\cat C_*}$.

In summary, semantic enrichment amounts to applying the annotation functor in
the category of elements. The expansion stage simply computes the functor. As an
aid to human interpretation, the contraction stage computes a new syntactic
expression for the expanded morphism, grouping together boxes that do not
correspond to morphisms in the ontology category.

\section{The view from data science} \label{sec:data-science-viewpoint}

Like the code it analyzes, our AI system is a means, not an end. Its impetus is
the transformation of science, currently under way, towards greater openness,
transparency, reproducibility, and collaboration. As part of this
transformation, data and machines will both come to play a more prominent role
in science. In this section, we describe the major themes of this evolution of
the scientific process and how we hope our work will contribute to it. We also
demonstrate our system on a realistic data analysis from the open science
community.

\subsection{Towards networked science}

Although the World Wide Web has already radically changed the dissemination of
scientific research, its potential as a universal medium for representing and
sharing scientific knowledge is only just beginning to be realized. A vast
library of scientific books and papers is now available online, accessible
instantaneously and throughout the world. That is a remarkable achievement,
accomplished in only a few decades. However, this endorsement must be qualified
in many respects. Scientific articles are accessible---but only to certain
people, due to the prevalence of academic paywalls. Even when articles are
freely available, the associated datasets, data analysis code, and supporting
software may not be. These research artifacts are, moreover, often not amenable
to systematic machine processing. In short, most scientific research is now on
the Web, but it may not be accessible, reproducible, or readily intelligible to
humans or machines.

A confluence of social and technological forces is pushing the scientific
community towards greater openness and interconnectivity. The open access
movement is gradually eroding paywalls \cite{piwowar2018}. The replication
crisis affecting several branches of science has prompted calls for stricter
standards about research transparency, especially when reporting data analysis
protocols \cite{pashler2012,munafo2017}. A crucial standard of transparency is
\emph{reproducibility}: the ability of researchers to duplicate the complete
data analysis of a previous study, from the raw data to the final statistical
inferences \cite{goodman2016}. Reproducibility demands that all relevant
datasets, analysis code, and supporting software be available---the same
requirements imposed by our system.

Another driving force is the growing size and complexity of scientific data.
Traditionally, the design, data collection, data analysis, and reporting for a
scientific experiment has been conducted entirely within a single research group
or laboratory. That is changing. Large-scale observational studies and
high-throughput measurement devices are producing ever larger and richer
datasets, making it more difficult for the people who collect the data to also
analyze it. Creating richer datasets also increases the potential gains from
data sharing and reuse. The FAIR Data Principles aim to simplify data reuse by
making datasets more ``FAIR'': findable, accessible, interoperable, and reusable
\cite{wilkinson2016}. Organizations like the Accelerated Cure Project for
Multiple Sclerosis and the Parkinson Progression Marker Initiative are creating
integrated online repositories of clinical, biological, and imaging data
\cite{marek2011}. In a related development, online platforms like Kaggle, Driven
Data, and DREAM Challenges are crowdsourcing data analysis through data science
competitions.

Science, then, seems to be headed towards a world where all the products of
scientific research, from datasets to code to published papers, are fully open,
online, and accessible. In the end, we think this outcome is inevitable, even if
it is delayed by incumbent interests and misaligned incentives. The consequences
of this new ``networked science'' are difficult to predict, but they could be
profound \cite{hey2009,nielsen2012}. We and others conjecture that new forms of
open, collaborative science, where humans and machines work together according
to their respective strengths, will accelerate the pace of scientific discovery.

An obstacle to realizing this vision is the lack of standardization and
interoperability in research artifacts. Researchers cannot efficiently share
knowledge, data, or code, and machines cannot effectively process it, if it is
not represented in formats that they readily understand. We aim to address one
aspect of this challenge by creating semantic representations of data science
code. We will say shortly what kind of networked science applications we hope
our system will enable. But first we describe more concretely one particular
model of networked science, the data science challenge, and a typical example of
the analysis code it produces.

\subsection{An example from networked science} \label{sec:dream}

As a more realistic example, in contrast to \cref{sec:example}, we examine a
data analysis conducted for a DREAM Challenge. DREAM Challenges address
scientific questions in systems biology and translational medicine by
crowdsourcing data analysis across the biomedical research community
\cite{stolovitzky2007,stolovitzky2016}. Under the challenge model, teams compete
to create the best statistical models according to metrics defined by the
challenge organizers. Rewards may include prize money and publication
opportunities. In some challenges, the initial competitive phase is followed by
a cooperative phase where the best performing teams collaborate to create an
improved model \cite[see, for example,][]{dream-mammography-2017,sieberts2016}.

The challenge we consider asks how well clinical and genetic covariates predict
patient response to anti-TNF treatment for rheumatoid arthritis
\cite{sieberts2016}. Of special interest is whether genetic biomarkers can serve
as a viable substitute for more obviously relevant clinical diagnostics. To
answer this question, each participant was asked to submit two models, one using
only genetic covariates and the other using any combination of clinical and
genetic covariates. After examining a wide range of models, the challenge
organizers and participants jointly concluded that the genetic covariates do not
meaningfully increase the predictive power beyond what is already contained in
the clinical covariates.

\begin{figure*}
  \centering
  \includegraphics[width=\textwidth]{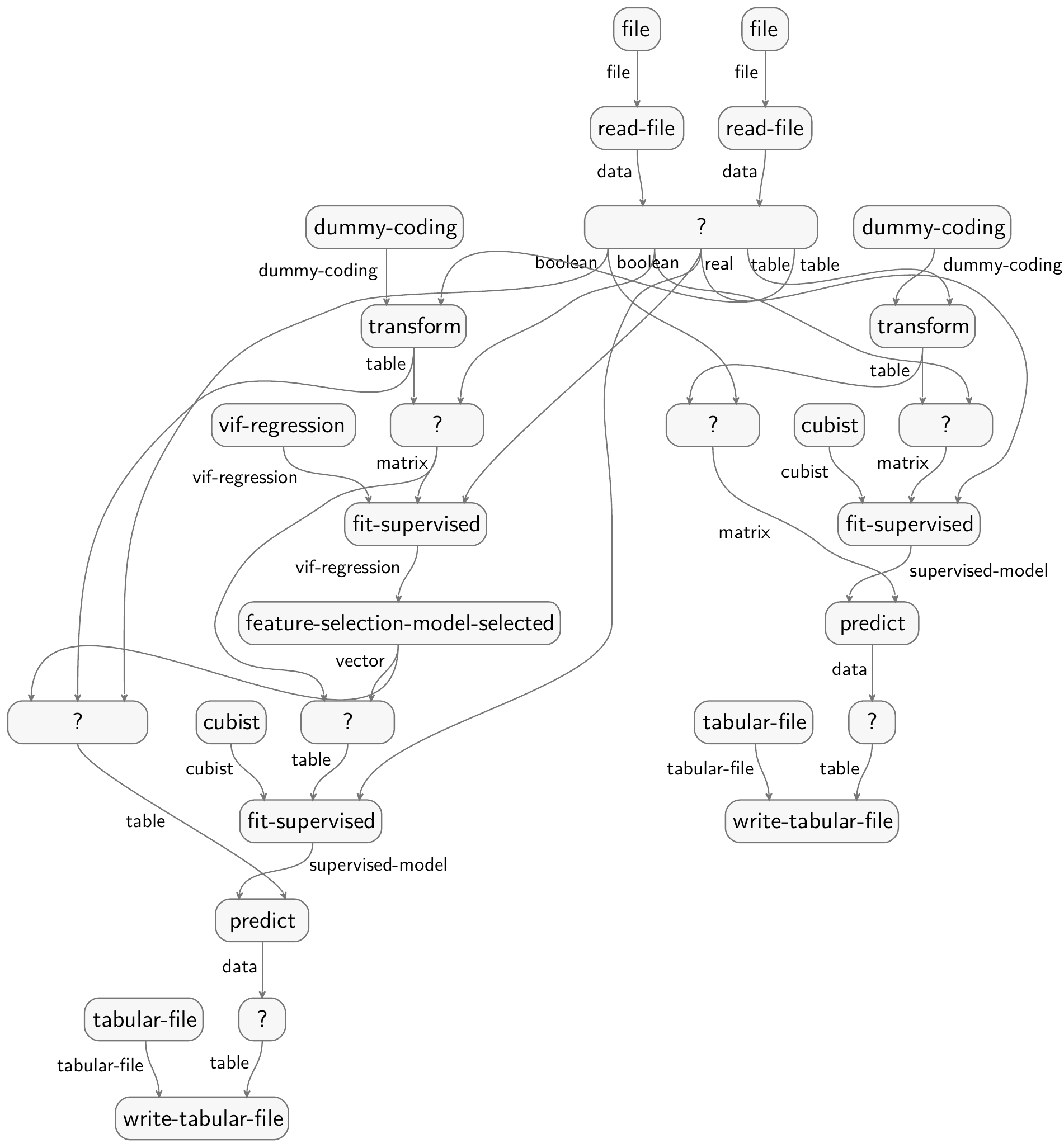}
  \caption{Semantic flow graph for two models from the Rheumatoid Arthritis
    DREAM Challenge (\cref{lst:dream-ra})}
  \label{fig:semantic-dream-ra}
\end{figure*}

We use our system to analyze the two models submitted by a top-ranking team
\cite{kramer2014}. The source code for the models, written in R, is shown in
\cref{lst:dream-ra}. It has been lightly modified for portability. The
corresponding semantic flow graph is shown in \cref{fig:semantic-dream-ra}. The
reader need not try to understand the code in any great detail. Indeed, we hope
that the semantic flow graph will be easier to comprehend than the code and
hence will be serve as an aid to humans as well as to machines. We grant,
however, that the current mode of presentation is far from ideal from the human
perspective.\footnote{We would prefer a web-based, interactive presentation,
  with the boxes and wires linked to descriptions from the ontology. That is,
  regrettably, outside the scope of this paper.}

\begin{listing}
  \begin{minted}[fontsize=\scriptsize,frame=leftline,rulecolor=\color{gray!50}]{r}
library("caret")
library("VIF")
library("Cubist")

merge.p.with.template <- function(p){
  template = read.csv("RAchallenge_Q1_final_template.csv")
  template$row = 1:nrow(template)
  template = template[,c(1,3)]
  
  ids = data.resp$IID[is.na(y)]
  p = data.frame(ID=ids, Response.deltaDAS=p)
  p = merge(template, p)
  p = p[order(p$row), ]
  p[,c(1,3)]
}

data = readRDS("pred.rds")
resp = readRDS("resp.rds")

# non-clinical model
data.resp = merge(data, resp[c("FID", "IID", "Response.deltaDAS")])
y = data.resp$Response.deltaDAS
y.training = y[!is.na(y)]

data.resp2 = data.resp[!(names(data.resp) %in% c("Response.deltaDAS", "FID", "IID"))]
dummy = predict(dummyVars(~., data=data.resp2), newdata=data.resp2)

dummy.training = dummy[!is.na(y),]
dummy.testing = dummy[is.na(y),]

v = vif(y.training, dummy.training, dw=5, w0=5, trace=F)
dummy.training.selected = as.data.frame(dummy.training[,v$select])
dummy.testing.selected = as.data.frame(dummy.testing[,v$select])

m1 = cubist(dummy.training.selected, y.training, committees=100)
p1 = predict(m1, newdata=dummy.testing.selected)

# clinical model
dummy = data.resp[c("baselineDAS", "Drug", "Age", "Gender", "Mtx")]
dummy = predict(dummyVars(~., data=dummy), newdata=dummy)
dummy.training = dummy[!is.na(y),]
dummy.testing = dummy[is.na(y), ]

m2 = cubist(dummy.training, y.training, committees=100)
p2 = predict(m2, newdata=dummy.testing)

## create csv files
p1.df = merge.p.with.template(p1)
p2.df = merge.p.with.template(p2)

write.csv(p1.df, quote=F, row.names=F, file="clinical_and_genetic.csv")
write.csv(p2.df, quote=F, row.names=F, file="clinical_only.csv")
  \end{minted}
  \caption{R source code for two models from the Rheumatoid Arthritis DREAM
    Challenge}
  \label{lst:dream-ra}
\end{listing}

The analysts fit two predictive models, the first including both genetic and
clinical covariates and the second including only clinical covariates. The
models correspond, respectively, to the first and second commented code blocks
and to the left and right branches of the semantic flow graph. Both models use
the Cubist regression algorithm \cite[\S 8.7]{kuhn2013}, a variant of random
forests based on M5 regression model trees \cite{wang1997}. Because the genetic
data is high-dimensional, the first model is constructed using a subset of the
genetic covariates, as determined by a variable selection algorithm called VIF
regression \cite{lin2011}. The linear regression model created by VIF regression
is used only for variable selection, not for prediction.

Most of the unlabeled nodes in \cref{fig:semantic-dream-ra}, including the wide
node at the top, refer to code for data preprocessing or transformation. It is a
commonplace among data scientists that such ``data munging'' is a crucial aspect
of data analysis. There is no fundamental obstacle to representing its
semantics; it so happens that the relevant portion of the Data Science Ontology
has not yet been developed. This situation illustrates another important point.
Our system does not need or expect the ontology to contain complete information
about the program's types and functions. It is designed to degrade gracefully,
producing useful partial results even in the face of missing annotations.

\subsection{Use cases and applications} \label{sec:applications}

Our system is a first step towards an AI assistant for networked, data-driven
science. We hope it will enable, or bring us closer to enabling, new
technologies that boost the efficiency of data scientists. These technologies
may operate at small scales, involving one or a small group of data scientists,
or at large scales, spanning a broader scientific community.

At the scale of individuals, we imagine an integrated development environment
(IDE) for data science that interacts with analysts at both syntactic and
semantic levels. Suppose a participant in the rheumatoid arthritis DREAM
Challenge fits a random forest regression, using the \texttt{randomForest}
package in R. Indeed, the analysts from \cref{sec:dream} report experimenting
with random forests, among other popular methods \cite{kramer2014}. By a simple
inference within the Data Science Ontology, the IDE recognizes random forests as
a tree-based ensemble method. It suggests the sister method Cubist and generates
R code invoking the \texttt{Cubist} package. Depending on their expertise, the
analysts may learn about new statistical methods or software packages
implementing them. Even expert users should benefit from the possibility of more
efficient experimentation.

As programmers, we are all prone to lapses of discipline in commenting our code.
Poorly documented code is difficult to comprehend at a glance. To supplement the
explanations written by fallible humans, an AI agent might translate our
semantic flow graphs into written descriptions, via natural language generation
\cite{gatt2018}. The playful R package \texttt{explainr} does exactly that for a
few R functions, in isolation \cite{parker2015}. A more comprehensive system
based on our work would span multiple languages and libraries and would document
both the individual steps and the high-level design of a data analysis.

New possibilities emerge at the scale of online platforms for collaborative data
science. Online platforms can host coordinated efforts to solve specific
scientific problems, under variations of the challenge model. They may also host
products of independent scientific experiments, serving as centralized
repositories of papers, data, and code. In both situations, statistical
meta-analysis is needed to aggregate the results of individual analyses or
studies \cite{gurevitch2018}. Today meta-analysis is a laborious and painstaking
process, conducted largely by hand. We hope to lay the groundwork for more
automated forms of meta-analysis.

Consider the challenge model again. Organizers typically want a panoramic view
of the participants' activity. We could straightforwardly generate a summary
report, given a corpus of semantic flow graphs corresponding to the submitted
analyses. In challenges of scientific interest, organizers tend to be interested
in more than simply what is the most predictive model. A recent DREAM Challenge,
for example, aims to determine which factors affect the progression of
amyotrophic lateral sclerosis (ALS), a fatal neurodegenerative disease with
heterogeneous progression timelines \cite{kueffner2018}. The organizers
stipulate that submitted predictive models may use only a limited number of
clinical features. Using consensus clustering \cite{monti2003}, the organizers
then aggregate feature sets across the submitted models to stratify the patients
into clinically meaningful subgroups. This and other forms of meta-analysis
could conceivably be simplified, or even automated, given sufficiently
expressive semantic representations of the models.

\section{Related work} \label{sec:related-work}

In this paper, we extend and refine our previous work on semantic
representations of data analyses \cite{patterson-ibm2017}. Compared to the
original work, we designed a new ontology language for modeling computer
programs, along with a new ontology about data science written in this language.
We also replaced our original, ad hoc procedure for creating the semantic flow
graph with the semantic enrichment algorithm, which is more flexible and rests
on a firmer mathematical foundation. We presented our current system as a
demonstration at IJCAI 2018 \cite{patterson-ijcai2018}.

We have been inspired by a constellation of ideas at the intersection of
artificial intelligence, program analysis, programming language theory, and
category theory. We now position our work in relation to these areas.

\subsection{Knowledge representation and program analysis}

The history of artificial intelligence is replete with interactions between
knowledge representation and computer program analysis. In the late 1980s and
early 1990s, automated planning and ruled-based expert systems featured in
``knowledge-based program analysis''
\cite{johnson1985,harandi1990,biggerstaff1994}. Other early systems were based
on description logic \cite{devanbu1991,welty2007} and graph parsing
\cite{wills1992}. Such projects are supposed to help software developers
maintain large codebases (exceeding, say, a million lines of code) in
specialized industrial domains like telecommunications.

Our research goals are less ambitious in scale but also, we hope, more
tractable. We focus on knowledge workers who write short, semantically rich
scripts, without the endless layers of abstraction found in large codebases. In
data science, the code tends to be much shorter, the control flow more linear,
and the underlying concepts better defined, than in large-scale industrial
software. Our methodology is accordingly quite different from that of the older
literature.

\subsection{Machine learning and program analysis}

Efforts are now underway to marry program analysis with machine learning.
Inspired by an analogy between natural languages and programming languages, AI
researchers are transporting successful techniques from natural language
processing (NLP), such as Markov models and recurrent neural networks, to
program analysis \cite{allamanis2018}. Most program models are based on
sequences of syntactic tokens, akin to sequences of words in natural language.
Some models use graphical program representations, bringing them closer to our
work. For example, a recent method called \texttt{inst2vec} (``instructions to
vectors''), inspired by \texttt{word2vec}, fits a skip-gram embedding of program
statements, using a notion of statement context which combines data flow and
control flow \cite{ben-nun2018}.

The logical and statistical paradigms of AI tend to exhibit different
performance characteristics and are therefore complementary, not competitive. In
the case of program analysis, our method delivers rich, precise, and
human-interpretable semantics, at the expense of significant human knowledge
engineering. Statistical methods scale better in terms of human effort and
degrade more gracefully in the face of incomplete information, but yield
semantics that are less precise and harder to interpret. In particular,
embedding methods like \texttt{inst2vec}\footnote{\texttt{inst2vec} also differs
  from our system by operating on LLVM's intermediate representation (IR), not
  the original code. This choice seems not to be viable for data science because
  Python and R do not have stable LLVM frontends, among other possible
  difficulties.} create dense vector representations of statements, whose
interpretations are defined only implicitly by their relation to other vectors.
The vectors are useful for downstream prediction tasks but are difficult to
interpret directly. Moreover, logical and statistical methods tend to understand
the slippery notion of ``semantics'' in fundamentally different ways. Vector
representations capture distributional information about how concepts are used
in practice, whereas ontologies express logical constraints on how concepts are
related to each other. Both kinds of information are useful and important. In
the future, we hope to investigate ways of integrating logical and statistical
information in semantic representations of data science code.

\subsection{Ontologies for data science}

There already exist several ontologies and schemas related to data science, such
as STATO, an OWL ontology about basic statistics \cite{gonzalez-beltran2016};
the Predictive Modeling Markup Language (PMML), an XML schema for data mining
models \cite{guazzelli2009}; and ML Schema, a schema for data mining and machine
learning workflows under development by a W3C community group
\cite{lawrynowicz2017}. What does the Data Science Ontology add to the
landscape? While we can certainly point to differences in content---STATO
focuses on classical statistics, especially hypothesis testing, whereas we are
equally interested in machine learning---we prefer to make a more general point,
applicable to all the ontologies that we know of.

Every ontology is, implicitly or explicitly, designed for some purpose. The
purpose of the Data Science Ontology is to define a universal language for
representing data science code. Previous ontologies were designed for different
purposes, and we cannot see any straightforward way to adapt them to ours. In
STATO, concepts representing statistical methods can have inputs and outputs,
but they are too imprecisely specified to map onto actual code, among other
difficulties. PMML is a purely static format, designed for serializing fitted
models. To successfully model computer programs, one must pay attention to the
special structure of programs. That is what we have tried to do with the Data
Science Ontology. This aspiration also drives our choice of ontology language.

\subsection{Ontology languages and programming languages}

We have designed an ontology language, Monocl, to model computer programs.
Although it is the medium of the Data Science Ontology, the ontology language is
conceptually independent of data science or any other computational domain.
Mathematically, it is founded on category theory and programming language
theory. References are given in \cref{sec:math}, where we develop the theory. We
hope that our project will advance an emerging paradigm of knowledge
representation based on category theory \cite{spivak2012,patterson-arxiv2017}.

Due in part to the influence of the Semantic Web, the most popular paradigm for
knowledge representation today is description logic, a family of computationally
tractable subsystems of first-order logic \cite{baader2007}. Why have we not
written the Data Science Ontology in a description logic, like the Semantic
Web's OWL? We do not claim this would be impossible. The Basic Formal Ontology,
expressible in OWL and underlying many biomedical ontologies, divides the world
into \emph{continuants} (persistent objects) and \emph{occurrents} (events and
processes) \cite{arp2015}. We might follow STATO in modeling data analyses, or
computational processes generally, as occurrents. This leads to some awkward
consequences, as occurrents are ascribed a spatiotemporal structure which
computer programs lack.

A more fundamental objection, independent of the Basic Formal Ontology, is that
there already exists a long mathematical tradition of modeling programs,
beginning nearly one hundred years ago with Alonzo Church's invention of the
lambda calculus. We follow a few threads in this tradition in
\cref{sec:math-motivation}. Our work very much belongs to it. To instead ignore
it, reinventing a programming model inside description logic, would be, at best,
an unnecessary duplication of effort. That said, we understand the value of
interoperability with existing systems. We are investigating ways to encode the
Data Science Ontology in OWL, possibly with some loss of fidelity.

\section{Conclusion} \label{sec:conclusion}

We have introduced an algorithm, supported by the Monocl ontology language and
the Data Science Ontology, for creating semantic representations of data science
code. We demonstrated the semantic enrichment algorithm on several examples,
pedagogical and practical, and we supplied it with a category-theoretic
mathematical foundation. We situated our project within a broader trend towards
more open, networked, and machine-driven science. We also suggested possible
applications to collaborative data science, at the small and large scales.

In future work, we plan to build on the suggestive examples presented in this
paper. We will develop methods for automated meta-analysis based on semantic
flow graphs and conduct a systematic empirical evaluation on a corpus of data
analyses. We also have ambitions to more faithfully represent the mathematical
and statistical structure of data science models. Our representation emphasizes
computational structure, but the most interesting applications require more. In
a similar vein, scientific applications require that statistical models be
connected with scientific concepts. Workers across the sciences, and especially
in biomedicine, are building large ontologies of scientific concepts.
Interoperation with existing domain-specific ontologies is therefore an
important research direction.

Only by a concerted community effort will the vision of machine-assisted data
science be realized. To that end, we have released as open source software our
Python and R program analysis tools, written in their respective languages, and
our semantic enrichment algorithm, written in Julia.\footnote{All source code is
  available on GitHub under the Apache 2.0 license
  \cite{patterson-pyflowgraph2018,patterson-rflowgraph2018,patterson-semanticflowgraph2018}.}
We are also crowdsourcing the further development of the Data Science
Ontology.\footnote{The Data Science Ontology is available on GitHub under the
  Creative Commons Attribution 4.0 license
  \cite{patterson-datascienceontology2018}.} We entreat the reader to join us in
the effort of bringing artificial intelligence to the practice of data science.

\printbibliography

\end{document}